\documentclass[sigconf]{acmart}
\AtBeginDocument{%
  }

\setcopyright{acmlicensed}
\copyrightyear{2018}
\acmYear{2018}
\acmDOI{XXXXXXX.XXXXXXX}
\acmConference[Conference acronym 'XX]{Make sure to enter the correct
  conference title from your rights confirmation email}{June 03--05,
  2018}{Woodstock, NY}
\acmISBN{978-1-4503-XXXX-X/2018/06}

\usepackage{multirow}
\usepackage{subcaption}

\begin{document}

\title[T-LLM]{T-LLM: Teaching Large Language Models to Forecast Time Series via Temporal Distillation}

\author{Suhan Guo}
\email{shguo@smail.nju.edu.cn}
\affiliation{
    \institution{State Key Laboratory for Novel Software Technology, Nanjing University}
    \institution{School of Artificial Intelligence, Nanjing University}
    \city{Nanjing}
    \state{Jiangsu}
    \country{China}
}

\author{Bingxu Wang}
\email{221240001@smail.nju.edu.cn}
\affiliation{
    \institution{State Key Laboratory for Novel Software Technology, Nanjing University}
    \institution{School of Artificial Intelligence, Nanjing University}
    \city{Nanjing}
    \state{Jiangsu}
    \country{China}
}

\author{Shaodan Zhang}
\email{shaodanzhang@smail.nju.edu.cn}
\affiliation{
    \institution{State Key Laboratory for Novel Software Technology, Nanjing University}
    \institution{School of Artificial Intelligence, Nanjing University}
    \city{Nanjing}
    \state{Jiangsu}
    \country{China}
}

\author{Furao Shen}
\authornote{Corresponding Author}
\email{frshen@nju.edu.cn}
\affiliation{
    \institution{State Key Laboratory for Novel Software Technology, Nanjing University}
    \institution{School of Artificial Intelligence, Nanjing University}
    \city{Nanjing}
    \state{Jiangsu}
    \country{China}
}

\renewcommand{\shortauthors}{Guo et al.}

\begin{abstract}
    Time series forecasting plays a critical role in decision-making across many real-world applications. Unlike data in vision and language domains, time series data is inherently tied to the evolution of underlying processes and can only accumulate as real-world time progresses, limiting the effectiveness of scale-driven pretraining alone. This time-bound constraint poses a challenge for enabling large language models (LLMs) to acquire forecasting capability, as existing approaches primarily rely on representation-level alignment or inference-time temporal modules rather than explicitly teaching forecasting behavior to the LLM. We propose T-LLM, a temporal distillation framework that equips general-purpose LLMs with time series forecasting capability by transferring predictive behavior from a lightweight temporal teacher during training. The teacher combines trend modeling and frequency-domain analysis to provide structured temporal supervision, and is removed entirely at inference, leaving the LLM as the sole forecasting model. Experiments on benchmark datasets and infectious disease forecasting tasks demonstrate that T-LLM consistently outperforms existing LLM-based forecasting methods under full-shot, few-shot, and zero-shot settings, while enabling a simple and efficient deployment pipeline.
\end{abstract}




\received{20 February 2007}
\received[revised]{12 March 2009}
\received[accepted]{5 June 2009}

\maketitle

% TL;DR
% T-LLM is a framework that teaches large language models to perform time series forecasting by distilling predictive behavior from lightweight temporal models during training.

\section{Introduction}

% Paragraph 1: What is forecasting and why does it matter
Time series forecasting is fundamental to decision-making in many domains, from finance to large-scale monitoring systems. In practice, forecasts are often required under uncertainty and limited historical data, making generalization a key challenge for modern forecasting models.

\begin{figure}[t]
    \centering
    \includegraphics[width=0.80\linewidth]{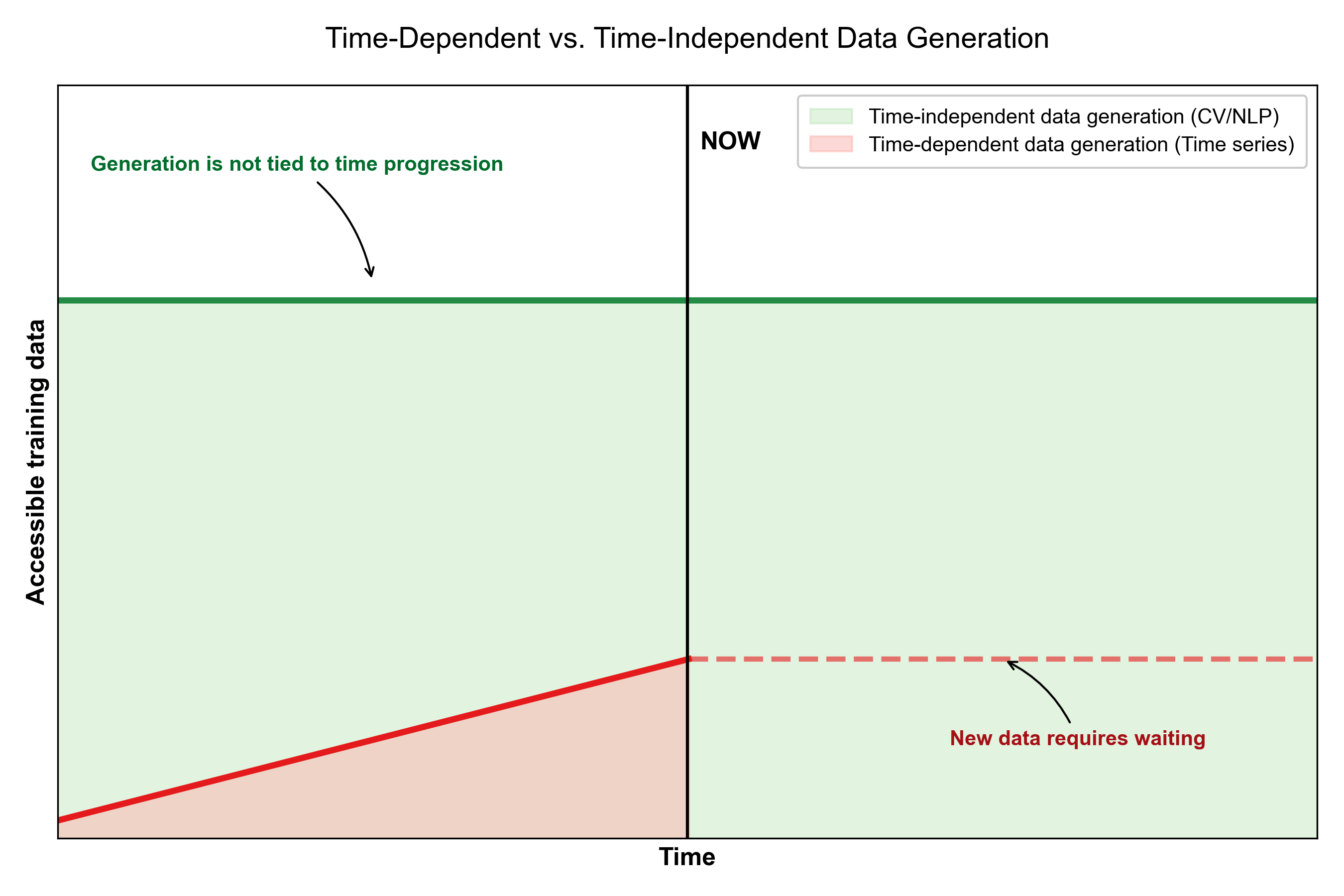}
    \caption{The Time-Bound Constraint. CV/NLP data can be collected independently of real-world time, whereas time-series data grows only as real processes evolve, creating a ``Future Wall'' for scale-driven pretraining.}
    \label{fig:current_challenge}
\end{figure}

% Paragraph 2: The current problem we identify
A particularly challenging forecasting setting arises in emergent and rapidly evolving domains, where models must operate in a zero-shot or near–zero-shot regime. Recent large-scale time-series models~\cite{jinTimeLLMTimeSeries2024,ZhouOneFitsAll2023NiPS,shi2025timemoe} demonstrate that strong generalization can be achieved through massive pretraining, establishing scale as an effective mechanism for acquiring forecasting capability. However, unlike image or language data, time series data are inherently constrained by real-world time: observations are generated sequentially, past data are fixed, and additional data can only be obtained by waiting. As a result, large-scale time-series pretraining is not only costly but fundamentally time-bound, making scale-driven strategies inherently exhaustive rather than extensible. This raises a complementary and largely unexplored question:
\textbf{can a general-purpose large language model be equipped with time series forecasting capability without relying on large-scale time-series pretraining?}

% version 1
% A particularly challenging forecasting setting arises in emergent and data-scarce domains, where models must operate in a zero-shot or near–zero-shot regime. Recent large-scale time-series models~\cite{jinTimeLLMTimeSeries2024,ZhouOneFitsAll2023NiPS,shi2025timemoe} demonstrate that strong generalization can be achieved through massive pretraining. However, this paradigm relies on scale as the primary mechanism for acquiring forecasting capability. As time-series data are finite and costly to curate at scale, strategies that rely primarily on ever-larger pretraining are inherently exhaustive rather than extensible. This raises a complementary and largely unexplored question: \textbf{can a general-purpose large language model be equipped with time series forecasting capability without relying on large-scale time-series pretraining?}

% Paragraph 3: What existing methods have done
Recent work on time series forecasting has explored several complementary directions. One line of research adapts large language models (LLM) to numerical time series by aligning representations or introducing auxiliary temporal branches, as exemplified by CALF~\cite{liuCALFAligningLlms2025} and TimeLLM~\cite{jinTimeLLMTimeSeries2024}. These methods aim to leverage the generalization capacity of LLMs while compensating for their lack of native temporal modeling through architectural alignment or prompt-based mechanisms. Separately, a large body of work has focused on improving temporal modeling directly through specialized architectures. Methods such as TSLANet~\cite{eldeleTSLANetRethinkingTransformers2024} incorporate spectral analysis to capture periodic behavior, while models like T3Time~\cite{chowdhuryT3TimeTriModalTime2025} introduce gating mechanisms to adaptively select relevant temporal patterns. These approaches demonstrate the effectiveness of explicitly modeling temporal structure but do not address how such knowledge can be transferred to LLM-based systems. More recently, large-scale foundation models for time series, such as Time-MoE~\cite{shi2025timemoe}, have shown strong zero-shot forecasting performance by pretraining on massive and diverse time-series corpora. These models represent a scale-driven approach to general-purpose forecasting, relying on extensive pretraining to internalize temporal regularities across domains.

% Paragraph 4: What is still lacking
Despite recent progress, existing approaches do not explicitly address how forecasting ability is acquired by an LLM. Alignment-based methods connect time-series representations to language models, but the LLM itself does not learn to forecast and remains dependent on external temporal modules. Foundation models, in contrast, do learn forecasting behavior, but only as an emergent consequence of large-scale exposure accumulated over real-world time. What remains missing is a training paradigm that treats forecasting as a transferable skill, one that can be taught to an LLM through structured temporal guidance, rather than waiting for future data to arrive in order to learn from it.

% Paragraph 5: Our solution and contributions
In this work, we address this gap by reframing LLM-based forecasting as a temporal distillation problem. Rather than treating cross-modal alignment as an end goal, we frame forecasting as a skill to be learned and train an LLM to internalize forecasting behavior by imitating the predictions of a lightweight temporal teacher. Our contributions can be summarized as follows:
\begin{itemize}
    % Zero-shot time series forecasting for emergent infectious diseases is typically addressed through large-scale foundation models trained on massive time-series corpora, an approach that is computationally expensive and difficult to reproduce or adapt in many research settings.
    \item We study a complementary alternative based on reverse distillation, enabling zero-shot forecasting by injecting explicit temporal pattern, such as trends and frequency-related behavior, into a pretrained LLM through lightweight supervision from a structured temporal teacher, without requiring large-scale time-series pretraining.
    
    % Existing LLM-based time series forecasting methods primarily rely on representation-level alignment between numerical time series and language models, but such alignment alone does not guarantee that the LLM acquires effective forecasting capability, often leaving auxiliary temporal modules necessary at inference time.
    \item We combine representation-level interaction with prediction-level supervision, explicitly training the LLM to reproduce the forecasts of a temporally specialized branch, thereby ensuring that forecasting capability is transferred into the LLM and enabling the temporal branch to be removed at inference.
    
    % Existing LLM-based forecasting frameworks provide limited guidance on what forms of temporal supervision are effective for teaching LLMs to forecast time series, particularly in capturing complementary temporal patterns such as trends and periodic behavior across different forecasting horizons.
    \item We design a lightweight temporal–spectral teacher that combines DLinear-based trend modeling with FFT-based frequency analysis, framed as structured supervision to expose the LLM to diverse temporal patterns relevant for both short- and long-horizon forecasting tasks.

    % TODO: add real-statistics
    \item We perform extensive empirical evaluations on benchmark time series datasets and infectious disease forecasting tasks, showing that reverse distillation consistently improves forecasting performance over existing LLM-based baselines in full-shot settings, while retaining strong generalization in zero-shot deployment.
\end{itemize}

\section{Related Works}

\subsection{Time Series Forecasting}

Time series forecasting has seen rapid progress driven by deep learning, with Transformer-based models achieving strong performance by modeling long-range dependencies and complex temporal interactions~\cite{nieTimeSeriesWorth2023,zhangCrossformerTransformerUtilizing2023,wuTimesNetTemporal2DVariation2023,zhouFEDformerFrequencyEnhanced2022}. Despite their effectiveness, these models often require large parameter budgets and substantial training data, limiting their applicability in data-constrained settings.

Recent studies have shown that lightweight forecasters can achieve competitive performance with significantly fewer parameters. Decomposition-based linear models such as DLinear explicitly separate trend and seasonal components to improve stability and efficiency~\cite{zengAreTransformersEffective2023}, while frequency-domain approaches capture periodic structure using compact spectral representations~\cite{caoSpectralTemporalGraph2020,yiFrequencydomainMLPsAre2023,eldeleTSLANetRethinkingTransformers2024}. Moreover, empirical evidence suggests that attention mechanisms are not always essential for forecasting accuracy~\cite{guo2025ramreplaceattentionmlp,zengAreTransformersEffective2023}, highlighting the cost-effectiveness of lightweight temporal models.

However, both heavyweight and lightweight forecasters typically rely on full-shot training and struggle in zero-shot or near–zero-shot scenarios. To address this limitation, recent work has explored adapting large language models (LLMs) for time series forecasting by reprogramming inputs or prompts~\cite{ZhouOneFitsAll2023NiPS,jinTimeLLMTimeSeries2024,sunTESTTextPrototype2024}. Large foundation models such as Time-MoE further demonstrate strong zero-shot performance through massive pretraining~\cite{shi2025timemoe}, but depend on extensive data and computational resources.

These developments motivate alternative training strategies that enable zero-shot forecasting without large-scale time series pretraining, particularly by leveraging efficient temporal models as sources of transferable forecasting knowledge.

\subsection{Cross-modal Training}

Cross-modal training has been widely studied as a way to transfer knowledge across heterogeneous data modalities. In the context of time series forecasting, however, our goal is not to perform representation alignment between language and temporal data. Instead, we are interested in introducing a different training paradigm for enabling forecasting with large language models. Existing work on cross-modal alignment nonetheless provides the closest related foundation for this line of research.

Recent approaches adapt LLMs to time series tasks through token reprogramming, prompt-based inputs, or feature-level alignment between temporal representations and language embeddings~\cite{jinTimeLLMTimeSeries2024,xuePromptCastNewPromptbased2024,panS2IPLLMSemanticSpace2024,liuTimeCMALLMempoweredMultivariate2025}. These methods leverage pretrained language representations for forecasting, but are often constrained by modality mismatch and static fusion between language and numerical signals.

Representative examples include CALF, which introduces a dual-branch framework that aligns temporal and textual representations through feature-level regularization~\cite{caoTEMPOPromptbasedGenerative2024}, and T3Time~\cite{chowdhuryT3TimeTriModalTime2025}, which proposes adaptive pooling and gating mechanisms to balance trend and periodic information in forecasting models~\cite{wooCoSTContrastiveLearning2022}. While effective within their respective formulations, these approaches primarily focus on representation alignment or architectural fusion, leaving open the question of how forecasting behavior itself can be transferred into an LLM through temporal supervision during training.

\begin{figure*}[t]
    \centering
    \includegraphics[width=0.85\linewidth]{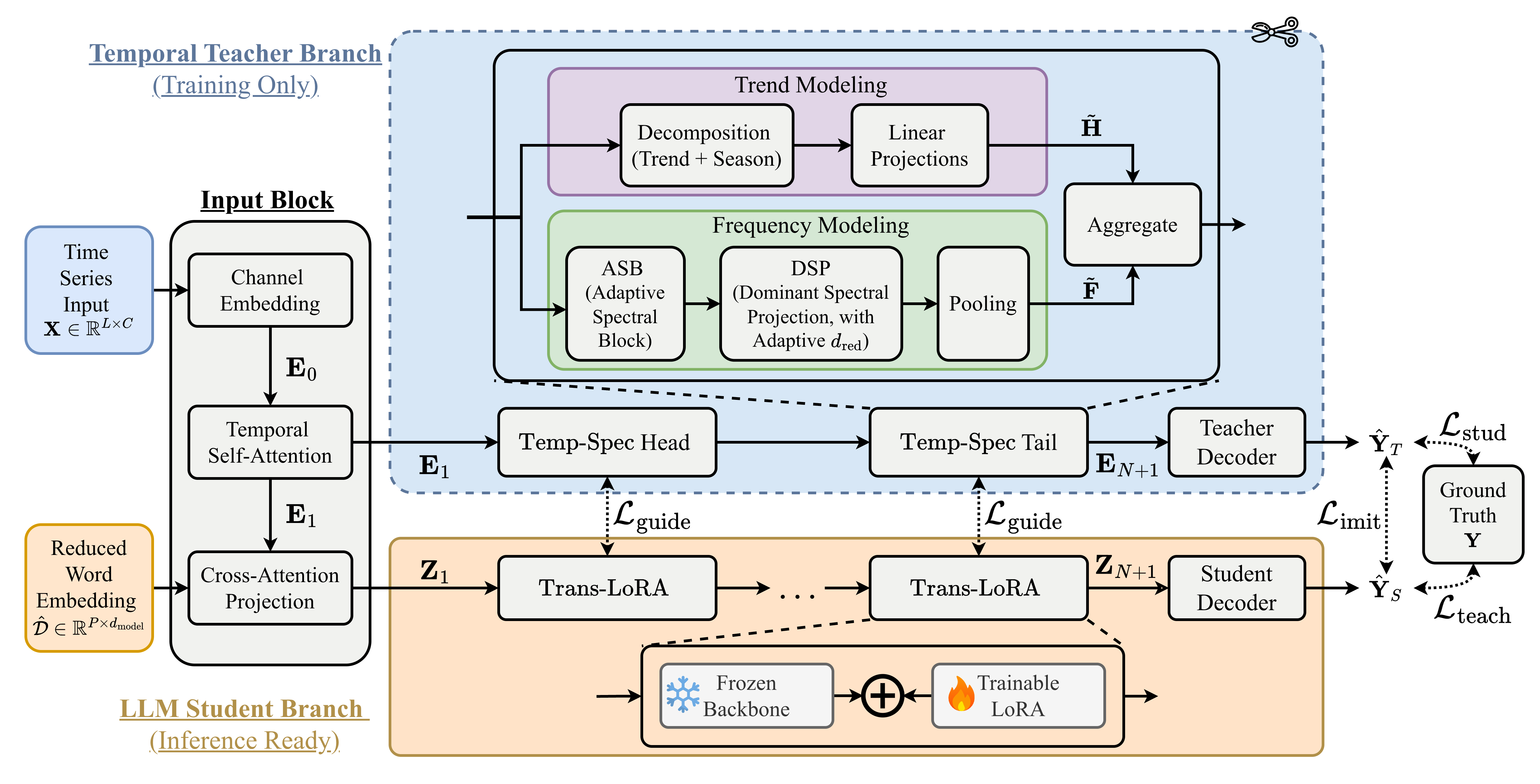}
    \caption{Overview of the proposed temporal distillation framework with a training-only temporal teacher and an inference-ready LLM student.}
    \Description{This figure illustrates the proposed T-LLM framework for time series forecasting. Given a multivariate time series input, an input encoding block produces shared representations that are fed into two branches during training: a temporal teacher and an LLM student. The temporal teacher is a lightweight model that captures trend and periodic structure using linear and spectral processing, and generates forecasts used for supervision. The LLM student is a pretrained transformer augmented with parameter-efficient adaptation modules and learns to internalize forecasting behavior through prediction-level and selective feature-level guidance from the teacher. The teacher branch is used only during training and is removed after distillation. At inference time, forecasting is performed solely by the trained LLM student and its decoder.}
    \label{fig:pipeline}
\end{figure*}

\section{Methods}
\subsection{Preliminary}
In multivariate time series forecasting, the objective is to predict future values based on a sequence of past observations. Let $\mathbf{X} = \{\mathbf{x}_1, \mathbf{x}_2, \dots, \mathbf{x}_L\} \in \mathbb{R}^{L \times C}$ denote an input time series of length L with C channels, and let $\hat{\mathbf{Y}} = \{\mathbf{x}_{L+1}, \mathbf{x}_{L+2}, \dots, \mathbf{x}_{L+T}\} \in \mathbb{R}^{T \times C}$ denote the prediction horizon of length $T$. We consider a forecasting model $\mathcal{F}(\cdot; \theta)$ parameterized by $\theta$, which maps historical observations to future predictions:
\begin{align}
    \mathrm{\hat{\mathbf{Y}}} = \mathcal{F}(\mathbf{X}; \theta).
\end{align}

\subsection{Overview of T-LLM}
As shown in Figure~\ref{fig:pipeline}, T-LLM is a framework designed to enable large LLMs to perform time series forecasting through temporal distillation. Rather than modifying the internal architecture of an LLM or relying on large-scale time-series pretraining, T-LLM teaches an LLM to forecast by supervising it with a lightweight, temporally specialized teacher model during training.

The framework consists of two main components: a temporal teacher branch and an LLM student branch. The temporal teacher is a compact forecasting model that captures key temporal patterns from numerical time series, while the LLM branch serves as a flexible function approximator that learns to reproduce the teacher’s forecasting behavior. During training, the two branches interact through representation-level information exchange and prediction-level supervision, allowing the LLM to internalize forecasting behavior guided by the teacher.

\subsection{Input Block}
Before feeding data into the teacher and student branches, we preprocess the input time series using embedding and attention-based projection layers, following the CALF framework.

We first define the multi-head attention operator as:
\begin{equation}
    \mathrm{MHA}(\mathbf{Q}, \mathbf{K}, \mathbf{V})
    := \mathrm{Softmax}\!\left(\frac{\mathbf{Q}\mathbf{K}^\top}{\sqrt{d_{\text{model}}}}\right)\mathbf{V},
\end{equation}
where $\mathbf{Q}, \mathbf{K}, \mathbf{V} \in \mathbb{R}^{n \times d_{\text{model}}}$ denote the query, key, and value matrices, respectively.

Let $\mathbf{X} \in \mathbb{R}^{L \times C}$ denote the input time series. Following CALF~\cite{liuCALFAligningLlms2025}, we treat channels as tokens and embed the input into the LLM feature space:
\begin{equation}
    \mathbf{E}_0 = \text{Embedding}(\mathbf{X}) \in \mathbb{R}^{C \times d_{\text{model}}}.
\end{equation}

To capture channel-wise temporal dependencies, we apply a self-attention layer:
\begin{align}
    \mathbf{Q}_T &= \mathbf{E}_0 \mathbf{W}_q,\quad
    \mathbf{K}_T = \mathbf{E}_0 \mathbf{W}_k,\quad
    \mathbf{V}_T = \mathbf{E}_0 \mathbf{W}_v,\\
    \mathbf{E}_1 &= \mathrm{MHA}(\mathbf{Q}_T, \mathbf{K}_T, \mathbf{V}_T),
\end{align}
where $\mathbf{W}_{\{q,k,v\}} \in \mathbb{R}^{d_{\text{model}} \times d_{\text{model}}}$ are learnable parameters. Since the student branch is a pretrained LLM, we further project temporal features into the textual embedding space using cross attention. We construct a compact word embedding dictionary $\hat{\mathcal{D}} \in \mathbb{R}^{P \times d_{\text{model}}}$ from the original embedding matrix $\mathcal{D} \in \mathbb{R}^{|\mathcal{A}| \times d_{\text{model}}}$, with $P \ll |\mathcal{A}|$. Cross attention is then applied as:
\begin{align}
    \mathbf{Q}_S &= \mathbf{E}_1 \mathbf{W}_q',\quad
    \mathbf{K}_S = \hat{\mathcal{D}} \mathbf{W}_k',\quad
    \mathbf{V}_S = \hat{\mathcal{D}} \mathbf{W}_v', \\
    \mathbf{Z}_1 &= \mathrm{MHA}(\mathbf{Q}_S, \mathbf{K}_S, \mathbf{V}_S).
\end{align}
Finally, $\mathbf{E}_1$ and $\mathbf{Z}_1$ are used as inputs to the temporal teacher branch and the LLM student branch, respectively.

\subsection{Temporal Teacher Branch}
The temporal teacher branch is designed to explicitly model temporal structure from numerical time series. It consists of lightweight components that capture both trend-related and frequency-related patterns. Details of each component are described in the following subsections.

\subsubsection{Trend Modeling}

The temporal teacher branch models long-term trend dynamics using a decomposition-based linear forecaster inspired by DLinear~\cite{zengAreTransformersEffective2023}. This component provides stable linear trend supervision for the LLM student, which is often difficult for LLMs to capture directly. Given the input representation $\mathbf{E}_1 \in \mathbb{R}^{C \times d_{\text{model}}}$, we first apply a moving-average-based decomposition operator that separates the input into a trend component and a residual (seasonal) component:
\begin{equation}
    \mathbf{H}_{\text{trend}} + \mathbf{H}_{\text{season}}=\mathbf{E}_1 ,
\end{equation}
where $\mathbf{H}_{\text{trend}}$ captures the coarse temporal trend and $\mathbf{H}_{\text{season}}$ represents the remaining variation.

Two independent linear projections are then applied to the decomposed components:
\begin{equation}
    \tilde{\mathbf{H}}_{\text{trend}} = \mathbf{H}_{\text{trend}} \mathbf{W}_{\text{trend}}, \quad
    \tilde{\mathbf{H}}_{\text{season}} = \mathbf{H}_{\text{season}} \mathbf{W}_{\text{season}},
\end{equation}
where $\mathbf{W}_{\text{trend}}, \mathbf{W}_{\text{season}} \in \mathbb{R}^{d_{\text{model}} \times d_{\text{model}}}$ are learnable linear mappings. The outputs are combined to produce the trend-based temporal representation:
\begin{equation}
    \tilde{\mathbf{H}} = \tilde{\mathbf{H}}_{\text{trend}} + \tilde{\mathbf{H}}_{\text{season}}.
\end{equation}
This decomposition-based linear modeling captures coarse temporal structure while remaining lightweight and stable during training.

\subsubsection{Frequency Modeling}

To complement trend-based temporal modeling, the temporal teacher branch incorporates frequency-domain analysis using an adaptive spectral processing block inspired by the Adaptive Spectral Block (ASB) in TSLANet~\cite{eldeleTSLANetRethinkingTransformers2024}. Frequency-domain filtering corresponds to circular convolution in the time domain, which offers a global receptive field and is particularly effective at capturing periodic patterns. This capability complements linear trend modeling, motivating the inclusion of adaptive spectral processing in the temporal teacher. We summarize this block as:
\begin{equation}
    \mathbf{F}_{\text{spec}} = \text{ASB}(\mathbf{E}_1).
\end{equation}
The details of this block can be found in Appendix section~\ref{app-sec:asb_block}.

% Dimension Reduction
In frequency modeling, spectral analysis is often applied along the raw sequence-length dimension. For example, TSLANet~\cite{eldeleTSLANetRethinkingTransformers2024} performs FFT directly over the temporal horizon $L$. In our framework, the sequence is first embedded into a latent space of dimension $d_{\text{model}}$, resulting in spectral analysis over a higher-dimensional representation. For real-valued inputs, retaining only the non-redundant Fourier coefficients yields a frequency-domain representation of length $d_{\text{FFT}}$, which can satisfy $d_{\text{FFT}} \gg L$. This places spectral analysis in a higher-dimensional space where informative frequency components are often sparse. In such settings, retaining the full spectrum may introduce redundant or weak responses.

Motivated by this observation, we apply spectral dimension reduction prior to pooling to retain dominant periodic structures. Specifically, we introduce a Dominant Spectral Projection (DSP) module that compresses the frequency dimension of the spectral representation:
\begin{equation}
    \tilde{\mathbf{F}}_{\text{spec}} = \text{DSP}(\mathbf{F}_{\text{spec}}) = \mathbf{F}_{\text{spec}} \mathbf{W}_{\text{spec}},
\end{equation}
where $\mathbf{W}_{\text{spec}} \in \mathbb{R}^{d_{\text{FFT}} \times d_{\text{red}}}$ is a learnable projection matrix and $d_{\text{red}} \ll d_{\text{FFT}}$.

% Adaptive Spectral Capacity Selection

In frequency-domain temporal modeling, the amount of spectral information required for effective forecasting depends strongly on the prediction horizon. Longer horizons demand richer periodic context to capture long-range seasonal structure, while shorter horizons benefit from compact representations that suppress spurious high-frequency responses. Using a fixed spectral capacity across forecasting tasks is therefore suboptimal. To address this, we introduce a horizon-conditioned spectral capacity schedule that regulates the amount of frequency information exposed to the LLM during distillation. Rather than treating spectral capacity as a tunable hyperparameter, we interpret it as horizon-aware capacity regularization on the temporal teacher. Specifically, let
\begin{equation}
    \mathcal{C} = \{(T_i, d_i)\}_{i=1}^{M},
\end{equation}
denote a predefined set of forecasting horizon–capacity pairs, where each $d_i$ specifies the maximum number of dominant frequency components allocated for horizon $T_i$. Given a forecasting horizon T, the reduced spectral dimension is selected as
\begin{equation}
    (T^*, d^*) = \arg\min_{(T_i, d_i)\in\mathcal{C}} |T - T_i|,\quad
    d_{\mathrm{red}}(T) = d^*.
\end{equation}
The DSP is then instantiated with
\begin{equation}
    \mathbf{W}_{\mathrm{spec}} \in \mathbb{R}^{d_{\mathrm{FFT}} \times d_{\mathrm{red}}(T)}.
\end{equation}
This discrete, horizon-conditioned capacity control avoids per-task tuning while regularizing the teacher’s supervision signal.

% Pooling
In contrast to prior spectral modeling approaches that reconstruct time-domain signals via inverse FFT, our temporal teacher uses pooling, proposed by T3Time~\cite{chowdhuryT3TimeTriModalTime2025}, to aggregate frequency-domain features. Since the teacher’s role is to provide compact supervision rather than signal reconstruction, pooling directly summarizes dominant spectral responses. We summarize the pooling as:
\begin{equation}
    \tilde{\mathbf{F}} = \text{Pooling}(\mathbf{F}_{\text{spec}}),
\end{equation}
with detailed description available at Appendix section~\ref{app-sec:pooling}.

\subsubsection{Trend–Periodic Aggregation}

The temporal teacher combines trend-based and frequency-based representations using a horizon-aware aggregation mechanism adapted from T3Time~\cite{chowdhuryT3TimeTriModalTime2025}. This module provides a simple, effective way to balance coarse trend signals and periodic structure based on the forecasting configuration. Since the aggregation strategy itself is not the focus of this work, we summarize it as:
\begin{equation}
    \mathbf{E}_2 = \text{Aggregate}(\tilde{\mathbf{H}}, \tilde{\mathbf{F}}),
\end{equation}
and refer interested readers to Appendix~\ref{app-sec:gating} for implementation details.

% Summary of temporal branch
In summary, the temporal teacher branch is defined as a stacked operator:
\begin{equation}
    \mathbf{E}_{n+1} = \text{Temp-Spec}(\mathbf{E}_n),
\end{equation}
where $\text{Temp-Spec}(\cdot)$ denotes a temporal–spectral processing block. In our implementation, the number of layers is set to $N=2$.

\subsection{LLM Student Branch}
The LLM student branch learns forecasting behavior under supervision from the temporal teacher. It is instantiated using a pretrained transformer-based language model and adapted through parameter-efficient fine-tuning~\cite{huLoRALowrankAdaptation2022}.

Formally, the student consists of transformer blocks augmented with low-rank adaptation, summarized as:
\begin{equation}
    \mathbf{Z}_{n+1} = \text{Trans-LoRA}(\mathbf{Z}_n),
\end{equation}
where $\text{Trans-LoRA}(\cdot)$ denotes a transformer block augmented with low-rank adaptation. We use up to $N=6$ layers in practice as GPT-2~\cite{radfordLanguageModelsAre2019} did. During training, the backbone parameters are frozen, and only the LoRA modules are optimized. 

% Output
The outputs of the two branches are mapped to forecasting targets using MLP-based decoders:
\begin{equation}
    \hat{\mathbf{Y}}_T = \text{Decoder}_T(\mathbf{E}_{N+1}), \;
    \hat{\mathbf{Y}}_S = \text{Decoder}_S(\mathbf{Z}_{N+1}),
\end{equation}
where $\hat{\mathbf{Y}}_T$ and $\hat{\mathbf{Y}}_S$ denote predictions from the temporal teacher and LLM student branches, respectively. At inference time, only $\hat{\mathbf{Y}}_S$ is used.

\subsection{Reverse Distillation}
We formulate training as a temporal distillation process, in which a lightweight temporal teacher provides forecasting supervision to an LLM student. The temporal teacher is optimized for numerical forecasting accuracy and defines the reference temporal behavior during training. The LLM student is trained to imitate the teacher’s predictions, receive coarse temporal guidance at selected layers, and maintain predictive stability through lightweight supervised calibration. Rather than aligning internal representations across modalities, our goal is to transfer forecasting behavior from a temporally specialized teacher into an LLM, while retaining the LLM as the sole inference-time model.

\paragraph{Training Objective.}
The overall loss is defined as:
\begin{equation}
\mathcal{L}
=
\mathcal{L}_{\text{teach}}
+ \lambda_1 \mathcal{L}_{\text{imit}}
+ \lambda_2 \mathcal{L}_{\text{guide}}
+ \lambda_3 \mathcal{L}_{\text{stud}},
\end{equation}
where $\lambda_1$, $\lambda_2$, and $\lambda_3$ balance imitation, temporal guidance, and student-side supervision.

\paragraph{Supervised and Imitation Losses.}
Both teacher and student predictions are supervised using standard forecasting losses:
\begin{equation}
\mathcal{L}_{\text{teach}} = \text{sim}(\hat{\mathbf{Y}}_T, \mathbf{Y}), \quad
\mathcal{L}_{\text{stud}} = \text{sim}(\hat{\mathbf{Y}}_S, \mathbf{Y}),
\end{equation}
where $\text{sim}(\cdot)$ is a similarity function such as the L1 loss. The teacher supervision ensures that the temporal branch learns accurate forecasting behavior, while student supervision regularizes the LLM prediction and prevents excessive deviation from the target forecasting space during training.

In addition, we introduce an imitation loss that encourages the student to reproduce the teacher’s predictions:
\begin{equation}
\mathcal{L}_{\text{imit}} = \text{sim}(\hat{\mathbf{Y}}_S, \hat{\mathbf{Y}}_T).
\end{equation}
This imitation objective serves as the primary mechanism for transferring forecasting behavior from the temporally specialized teacher to the LLM student. However, relying on imitation alone can be unstable when teacher signals are imperfect or when the student lags significantly behind the teacher. Combining imitation with direct supervision on the student prediction provides a stabilizing effect, accelerating convergence and ensuring that the student learns both to match the teacher and to remain anchored to the ground-truth forecasting objective.

\paragraph{Temporal Guidance Loss.}
To provide structural temporal cues without enforcing full representation alignment, we apply guidance at selected depths of the network:
\begin{equation}
\mathcal{L}_{\text{guide}}
=
\sum_{k \in \mathcal{K}}
\omega_k \,
\text{sim}\!\left(
\psi_k^S(\mathbf{Z}),
\psi_k^T(\mathbf{E})
\right),
\end{equation}
where $\mathcal{K}=\{2,3\}$ indexes early and late layers, $\psi_k^S(\cdot)$ and $\psi_k^T(\cdot)$ are projection heads, and $\omega_k$ controls their contributions. Unlike full layer-wise guidance, which constrains representations throughout the network, this selective guidance strategy injects temporal structure only at key stages of feature formation and output abstraction. Early-layer guidance helps ground low-level representations in temporal structure, while late-layer guidance aligns high-level forecasting semantics. This design is motivated by attention decay analysis (Appendix~\ref{app-sec:att_decay}), which shows that intermediate layers contribute less to effective temporal modeling.

\paragraph{Training and Inference.}
Training proceeds jointly over the teacher and student branches, with early stopping determined by convergence of the temporal teacher to ensure a stable supervision signal. This prevents the student from imitating overfitted teacher behavior. After training, the temporal teacher and all guidance-related components are removed, and forecasting is performed solely by the LLM student.

\section{Experiments}
To evaluate the effectiveness of T-LLM, we conduct comprehensive experiments across multiple time series forecasting settings, including long-term forecasting, short-term forecasting, and few/zero-shot learning. In addition, we analyze computational efficiency to assess the practicality of the proposed training paradigm.

\subsection{Baselines}
% LLM-based: CALF(Textual) CALF(Temporal) TimeLLM GPT4TS T3Time
%% TimeCMA UniTime
% Transformer-based: PatchTST iTransformer Crossformer ETSformer FEDformer Autoformer
% CNN-based: TCN MICN TimesNet
% MLP-based: DLinear TiDE N-HiTS N-BEATS

% TODO: Check if baselines are consistent with the results reported
We compare T-LLM against a diverse set of recent time series forecaster: 1) \textbf{LLM-based models} include CALF~\cite{liuCALFAligningLlms2025}, TimeLLM~\cite{jinTimeLLMTimeSeries2024}, GPT4TS~\cite{ZhouOneFitsAll2023NiPS}, and UniTime~\cite{liuUniTimeLanguageempoweredUnified2024}, which adapt pretrained language models to time series forecasting. For CALF, we report results from its textual prediction branch, which reflects the alignment-based training paradigm most closely related to our method. 2) \textbf{Transformer-based models} include PatchTST~\cite{nieTimeSeriesWorth2023}, iTransformer~\cite{liuITransformerInvertedTransformers2024}, %Crossformer~\cite{zhangCrossformerTransformerUtilizing2023}, 
and FEDformer~\cite{zhouFEDformerFrequencyEnhanced2022}. 3) \textbf{CNN-based models} include TCN~\cite{baiEmpiricalEvaluationGeneric2018}, MICN~\cite{wangMICNMultiscaleLocal2023}, and TimesNet~\cite{wuTimesNetTemporal2DVariation2023}. 4) \textbf{MLP-based models} include DLinear~\cite{zengAreTransformersEffective2023} and TiDE~\cite{dasLongtermForecastingTiDE2023}. For short-term forecasting tasks, we additionally include N-HiTS~\cite{challuNHITSNeuralHierarchical2023} and N-BEATS~\cite{oreshkinNBEATSNeuralBasis2020}, following standard evaluation practice.

% tab:longterm
\begin{table*}[]
\centering
\caption{Multivariate forecasting results. All results are averaged from four different forecasting horizons: $H \in \{96, 192, 336, 720\}$ for the input sequence length $96$. \textbf{Bold}: the best, \underline{underline}: the second best.}
\label{tab:longterm}
\resizebox{\textwidth}{!}{\begin{tabular}{@{}c|cc|cc|cc|cc|cc|cc|cc|cc|cc|cc|cc@{}}
\toprule
Models & \multicolumn{2}{c|}{T-LLM} & \multicolumn{2}{c|}{CALF} & \multicolumn{2}{c|}{TimeLLM} & \multicolumn{2}{c|}{GPT4TS} & \multicolumn{2}{c|}{UniTime} & \multicolumn{2}{c|}{PatchTST} & \multicolumn{2}{c|}{iTransformer} & \multicolumn{2}{c|}{FEDformer} & \multicolumn{2}{c|}{TimesNet} & \multicolumn{2}{c|}{DLinear} & \multicolumn{2}{c}{TiDE} \\ \midrule
Metric & MSE & MAE & MSE & MAE & MSE & MAE & MSE & MAE & MSE & MAE & MSE & MAE & MSE & MAE & MSE & MAE & MSE & MAE & MSE & MAE & MSE & MAE \\ \midrule
ETTm1 & 0.397 & \textbf{0.391} & 0.399 & \underline{0.392} & 0.41 & 0.409 & \underline{0.389} & 0.397 & \textbf{0.385} & 0.399 & 0.392 & 0.402 & 0.407 & 0.411 & 0.448 & 0.452 & 0.400 & 0.406 & 0.403 & 0.407 & 0.419 & 0.419 \\
ETTm2 & \textbf{0.278} & \textbf{0.319} & \underline{0.280} & \underline{0.320} & 0.296 & 0.340 & 0.285 & 0.331 & 0.293 & 0.334 & 0.285 & 0.328 & 0.291 & 0.335 & 0.305 & 0.349 & 0.291 & 0.333 & 0.350 & 0.401 & 0.358 & 0.404 \\
ETTh1 & \underline{0.441} & \textbf{0.433} & \underline{0.441} & \textbf{0.433} & 0.460 & 0.449 & 0.447 & \underline{0.436} & 0.442 & 0.448 & 0.463 & 0.449 & 0.455 & 0.448 & \textbf{0.440} & 0.460 & 0.458 & 0.450 & 0.456 & 0.452 & 0.541 & 0.507 \\
ETTh2 & \textbf{0.374} & \textbf{0.396} & \underline{0.375} & \underline{0.397} & 0.389 & 0.408 & 0.381 & 0.408 & 0.378 & 0.403 & 0.395 & 0.414 & 0.381 & 0.405 & 0.437 & 0.449 & 0.414 & 0.427 & 0.559 & 0.515 & 0.611 & 0.550 \\
weather & \textbf{0.246} & \textbf{0.271} & \underline{0.252} & \underline{0.276} & 0.274 & 0.290 & 0.264 & 0.284 & 0.253 & 0.276 & 0.257 & 0.280 & 0.257 & 0.279 & 0.309 & 0.360 & 0.259 & 0.287 & 0.265 & 0.317 & 0.271 & 0.320 \\
ECL & \textbf{0.174} & \textbf{0.265} & 0.181 & \underline{0.268} & 0.223 & 0.309 & 0.205 & 0.290 & 0.216 & 0.306 & 0.207 & 0.289 & \underline{0.178} & 0.270 & 0.214 & 0.327 & 0.192 & 0.295 & 0.212 & 0.300 & 0.251 & 0.344 \\
traffic & \underline{0.437} & \underline{0.287} & 0.448 & 0.294 & 0.541 & 0.358 & 0.488 & 0.317 & - & - & 0.481 & 0.304 & \textbf{0.428} & \textbf{0.282} & 0.610 & 0.376 & 0.620 & 0.336 & 0.625 & 0.383 & 0.760 & 0.473 \\ \midrule
\textbf{1st Count} & \textbf{4} & \textbf{6} & \textbf{0} & \textbf{1} & \textbf{0} & \textbf{0} & \textbf{0} & \textbf{0} & \textbf{1} & \textbf{0} & \textbf{0} & \textbf{0} & \textbf{1} & \textbf{1} & \textbf{1} & \textbf{0} & \textbf{0} & \textbf{0} & \textbf{0} & \textbf{0} & \textbf{0} & \textbf{0} \\ \bottomrule
\end{tabular}}
\end{table*}

% Table: shortterm
\begin{table*}[t]
\centering
\caption{Short-term forecasting performance on M4 with input lengths of 12–96 and horizons of 6–48.}
\label{tab:shortterm}
\resizebox{0.70\textwidth}{!}{\begin{tabular}{@{}cccccccccccc@{}}
\toprule
\multicolumn{2}{l}{Models} & T-LLM & CALF & TimeLLM & GPT4TS & PatchTST & TimesNet & TCN & N-HiTS & N-BEATS & DLinear \\ \midrule
\multirow{3}{*}{\rotatebox[origin=c]{90}{Yearly}} & SMAPE & \underline{13.362} & \textbf{13.343} & 13.419 & 13.531 & 13.477 & 13.387 & 14.920 & 13.418 & 13.436 & 16.965 \\
 & MASE & \textbf{2.982} & 3.033 & 3.005 & 3.015 & 3.019 & \underline{2.996} & 3.364 & 3.045 & 3.043 & 4.283 \\
 & OWA & \textbf{0.786} & 0.790 & 0.789 & 0.793 & 0.792 & \underline{0.786} & 0.880 & 0.793 & 0.794 & 1.058 \\ \midrule
\multirow{3}{*}{\rotatebox[origin=c]{90}{Quarterly}} & SMAPE & \textbf{10.030} & \underline{10.038} & 10.110 & 10.177 & 10.380 & 10.100 & 11.122 & 10.202 & 10.124 & 12.145 \\
 & MASE & \textbf{1.165} & \underline{1.168} & 1.178 & 1.194 & 1.233 & 1.182 & 1.360 & 1.194 & 1.169 & 1.520 \\
 & OWA & \textbf{0.880} & \underline{0.882} & 0.889 & 0.898 & 0.921 & 0.890 & 1.001 & 0.899 & 0.886 & 1.106 \\ \midrule
\multirow{3}{*}{\rotatebox[origin=c]{90}{Monthly}} & SMAPE & \textbf{12.634} & \underline{12.644} & 12.980 & 12.894 & 12.959 & 12.679 & 15.626 & 12.791 & 12.677 & 13.514 \\
 & MASE & \textbf{0.923} & \underline{0.927} & 0.963 & 0.956 & 0.970 & 0.933 & 1.274 & 0.969 & 0.937 & 1.037 \\
 & OWA & \textbf{0.872} & \underline{0.874} & 0.903 & 0.897 & 0.905 & 0.878 & 1.141 & 0.899 & 0.880 & 0.956 \\ \midrule
\multirow{3}{*}{\rotatebox[origin=c]{90}{Others}} & SMAPE & \underline{4.862} & 5.007 & \textbf{4.795} & 4.940 & 4.952 & 4.891 & 7.186 & 5.061 & 4.925 & 6.709 \\
 & MASE & \underline{3.187} & 3.220 & \textbf{3.178} & 3.228 & 3.347 & 3.302 & 4.677 & 3.216 & 3.391 & 4.953 \\
 & OWA & \underline{1.014} & 1.035 & \textbf{1.006} & 1.029 & 1.049 & 1.035 & 1.494 & 1.040 & 1.053 & 1.487 \\ 
 \midrule
 \multicolumn{2}{c}{\textbf{1st Count}} & \textbf{8} & \textbf{1} & \textbf{3} & \textbf{0} & \textbf{0} & \textbf{0} & \textbf{0} & \textbf{0} & \textbf{0} & \textbf{0} \\
 \bottomrule
\end{tabular}}
\end{table*}

\subsection{Implementation Details}
The LLM student branch is built on a pretrained GPT-2 model~\cite{radfordLanguageModelsAre2019}, where the first six transformer layers are used as the backbone. Optimization is performed using the Adam optimizer, with a learning rate of $0.0005$. For the total loss function, we set the hyper-parameters $\lambda_1 = 1.0, \lambda_2 = 0.01, \lambda_3 =1.0$. For long-term forecasting tasks, we use an L1 loss for supervised training on ETT datasets and smooth L1 loss for the remaining datasets, following prior practice. For short-term forecasting, we adopt SMAPE for supervised loss, MASE for prediction consistency, and smooth L1 loss for feature-level guidance. All experiments are conducted on a machine equipped with an NVIDIA GeForce RTX 4090 GPU (24 GB memory), running Ubuntu 22.04. Implementations are based on PyTorch 2.7.1 with Python 3.12.12.

\subsection{Long-term Forecasting}
\subsubsection{Setups}
We evaluate our proposed T-LLM on 7 real-world datasets: ETT (4 subsets), Traffic, Electricity, and Weather~\cite{zhouInformerEfficientTransformer2021, wuAutoformerDecompositionTransformers2021} for multivariate forecasting models. The input time series length T is fixed as $96$ for a fair comparison, and we adopt four distinct prediction horizons $\{96, 192, 336, 720\}$. Further details on the datasets can be found in Appendix~\ref{app-sec:dset}. Consistent with prior works~\cite{liuCALFAligningLlms2025}, the Mean Square Error (MSE) and Mean Absolute Error (MAE) are chosen as evaluation metrics.

\subsubsection{Results}

Table~\ref{tab:longterm} reports long-term forecasting performance across multiple benchmark datasets. T-LLM achieves the best or second-best performance across a majority of datasets and forecasting horizons, yielding the highest number of SOTA results among all compared methods. This indicates that lightweight temporal supervision with reverse distillation is more effective than alignment-based training for transferring forecasting behavior into the LLM.

Compared with heavier forecasting frameworks such as UniTime~\cite{liuUniTimeLanguageempoweredUnified2024}, T-LLM adopts a substantially lighter design, making training feasible on large-scale datasets such as Traffic. While models with extensive architectural augmentation can achieve strong accuracy, they often incur significantly higher computational and memory costs. For example, UniTime was not evaluated on Traffic in its original work, and we were unable to train it under our hardware constraints, highlighting a practical trade-off between accuracy and scalability.

Overall, these results emphasize that our contribution lies in the training paradigm rather than a specific model instantiation. Reverse distillation shows that forecasting behavior can be transferred into an LLM using lightweight temporal supervision, offering a complementary alternative to scale-driven approaches under time-bound data generation.

\subsection{Short-term Forecasting}
\subsubsection{Setups}
We evaluate our method on the M4 dataset~\cite{makridakisM4CompetitionResults2018}, which consists of univariate time series from marketing and business domains with yearly, quarterly, and monthly sampling frequencies. The forecasting horizons are relatively short, ranging from $6$ to $48$ steps, and the input window length is set to twice the corresponding prediction horizon. Following standard practice, performance is assessed using symmetric mean absolute percentage error (SMAPE), mean absolute scaled error (MASE), and the overall weighted average (OWA).

\subsubsection{Results}
Table \ref{tab:shortterm} reports short-term forecasting results on the M4 dataset. T-LLM consistently achieves the best or second-best performance across all evaluation metrics, outperforming most baseline models. In particular, CALF serves as a strong competitor and frequently attains the second-best results, confirming the effectiveness of the CALF framework for short-term forecasting. Compared to CALF, T-LLM further improves performance by transferring forecasting behavior from the temporal teacher to the LLM branch through reverse distillation.

\subsection{Few/Zero-shot Learning}
Few-shot and zero-shot settings provide a practical evaluation of model generalization when training data are limited or unavailable. We evaluate our method under both settings by restricting the amount of training data to 10\% in few-shot learning and transferring a trained model across datasets in zero-shot learning. 

% tab:Fewshot
\begin{table}[]
\centering
\caption{Few-shot forecasting results on 10\% training data. Input sequence is set to 96 for all benchmark datasets.}
\label{tab:fewshot}
\resizebox{0.48\textwidth}{!}{\begin{tabular}{@{}cc|cc|cc|cc|cc@{}}
\toprule
\multirow{2}{*}{} & Model & \multicolumn{2}{c|}{T-LLM} & \multicolumn{2}{c|}{CALF} & \multicolumn{2}{c|}{Time-LLM} & \multicolumn{2}{c}{GPT4TS} \\ \cmidrule(l){2-10} 
 & Pred Len & MSE & MAE & MSE & MAE & MSE & MAE & MSE & MAE \\ \midrule
\multirow{5}{*}{\rotatebox[origin=c]{90}{ETTh1}} & 96 & \textbf{0.396} & \underline{0.412} & 0.406 & 0.420 & 0.446 & 0.438 & \underline{0.398} & \textbf{0.405} \\
 & 192 & \textbf{0.446} & \underline{0.439} & 0.453 & 0.444 & 0.497 & 0.465 & \underline{0.448} & \textbf{0.433} \\
 & 336 & \textbf{0.486} & \underline{0.458} & 0.496 & 0.467 & 0.539 & 0.493 & \underline{0.496} & \textbf{0.456} \\
 & 720 & \textbf{0.498} & \textbf{0.479} & 0.527 & 0.499 & 0.536 & 0.507 & \underline{0.504} & 0.480 \\
 & Avg & \textbf{0.456} & \underline{0.447} & 0.470 & 0.457 & 0.504 & 0.476 & \underline{0.461} & \textbf{0.443} \\ \midrule
\multirow{5}{*}{\rotatebox[origin=c]{90}{ETTh2}} & 96 & \textbf{0.296} & \textbf{0.343} & 0.305 & 0.354 & 0.307 & 0.353 & \underline{0.303} & \underline{0.353} \\
 & 192 & \textbf{0.377} & \textbf{0.392} & 0.383 & \underline{0.394} & 0.395 & 0.406 & \underline{0.381} & 0.401 \\
 & 336 & 0.429 & \underline{0.430} & \underline{0.427} & \textbf{0.430} & 0.432 & 0.436 & \textbf{0.420} & 0.432 \\
 & 720 & \textbf{0.418} & \textbf{0.438} & \underline{0.422} & \underline{0.440} & 0.436 & 0.450 & 0.435 & 0.451 \\
 & Avg & \textbf{0.380} & \textbf{0.401} & \underline{0.384} & \underline{0.405} & 0.392 & 0.411 & 0.385 & 0.409 \\ \midrule
\multirow{5}{*}{\rotatebox[origin=c]{90}{ETTm1}} & 96 & \textbf{0.336} & \textbf{0.359} & 0.353 & 0.376 & 0.428 & 0.415 & \underline{0.349} & \underline{0.372} \\
 & 192 & \textbf{0.378} & \textbf{0.382} & 0.396 & 0.398 & 0.457 & 0.430 & \underline{0.388} & \underline{0.391} \\
 & 336 & \textbf{0.415} & \textbf{0.404} & \underline{0.420} & 0.413 & 0.487 & 0.446 & 0.422 & \underline{0.412} \\
 & 720 & \underline{0.491} & \underline{0.451} & 0.498 & 0.455 & 0.531 & 0.473 & \textbf{0.478} & \textbf{0.442} \\
 & Avg & \textbf{0.405} & \textbf{0.399} & 0.417 & 0.410 & 0.476 & 0.441 & \underline{0.409} & \underline{0.404} \\ \midrule
\multirow{5}{*}{\rotatebox[origin=c]{90}{ETTm2}} & 96 & 0.180 & \underline{0.260} & \textbf{0.179} & \textbf{0.259} & 0.193 & 0.277 & \underline{0.180} & 0.263 \\
 & 192 & \underline{0.246} & \textbf{0.302} & 0.247 & \underline{0.304} & 0.260 & 0.321 & \textbf{0.245} & 0.305 \\
 & 336 & 0.308 & \textbf{0.341} & \underline{0.307} & \underline{0.342} & 0.319 & 0.355 & \textbf{0.306} & 0.345 \\
 & 720 & \textbf{0.404} & \textbf{0.397} & \underline{0.406} & \underline{0.398} & 0.418 & 0.409 & 0.407 & 0.402 \\
 & Avg & 0.285 & \textbf{0.325} & \underline{0.285} & \underline{0.325} & 0.298 & 0.341 & \textbf{0.284} & 0.329 \\ \midrule
\multicolumn{2}{c|}{\textbf{1st Count}} & \textbf{14} & \textbf{13} & \textbf{1} & \textbf{2} & \textbf{0} & \textbf{0} & \textbf{5} & \textbf{5} \\ \bottomrule
\end{tabular}}
\end{table}

\subsubsection{Few-shot Learning}

Table~\ref{tab:fewshot} reports few-shot forecasting performance when models are trained on only 10\% of the training data. Across ETT benchmarks, T-LLM achieves strong and often SOTA performance under this limited-data regime, obtaining the best results on a majority of prediction lengths and datasets, particularly for short and medium horizons. These results indicate that temporal distillation enables the LLM to effectively leverage scarce supervision and maintain robust forecasting accuracy. We adopt a unified few-shot setting across all prediction horizons, rather than the protocol used in CALF. For long prediction lengths (e.g., 720), the CALF setting yields extremely limited effective training samples, leading to unstable optimization. By fixing the input sequence length and using a consistent 10\% training split, we ensure a fair and practical evaluation across horizons, under which T-LLM consistently matches or outperforms existing LLM-based baselines.

\begin{table}[t]
\centering
\caption{Zero-shot performance on ETT datasets. “$A \rightarrow B$” indicates that models trained on the dataset A are evaluated on dataset B.}
\label{tab:zero-shot}
\resizebox{0.48\textwidth}{!}{\begin{tabular}{@{}ccccccccc@{}}
\toprule
 & \multicolumn{2}{c}{h1$\rightarrow$m1} & \multicolumn{2}{c}{h1$\rightarrow$m2} & \multicolumn{2}{c}{h2$\rightarrow$m1} & \multicolumn{2}{c}{h2$\rightarrow$m2} \\ \cmidrule(lr){2-3} \cmidrule(lr){4-5} \cmidrule(lr){6-7} \cmidrule(lr){8-9}
 & MSE & MAE & MSE & MAE & MSE & MAE & MSE & MAE \\ \midrule
TiDE & 0.774 & 0.574 & \textbf{0.314} & \textbf{0.355} & 0.841 & 0.590 & \underline{0.321} & 0.364 \\
DLinear & \underline{0.760} & 0.577 & 0.399 & 0.439 & 0.778 & 0.594 & 0.496 & 0.496 \\
MICN & 1.439 & 0.780 & 2.428 & 1.236 & \underline{0.764} & 0.601 & 0.527 & 0.519 \\
TimesNet & 0.794 & 0.575 & 0.339 & 0.370 & 1.286 & 0.705 & 0.361 & 0.390 \\
FEDformer & 0.765 & 0.588 & 0.357 & 0.403 & \textbf{0.741} & \underline{0.588} & 0.365 & 0.405 \\
% Crossformer & 0.999 & 0.736 & 1.120 & 0.789 & 1.195 & 0.711 & 2.043 & 1.124 \\
PatchTST & 0.894 & 0.610 & 0.318 & 0.362 & 0.871 & 0.596 & 0.420 & 0.433 \\
GPT4TS & 0.798 & 0.574 & 0.317 & 0.359 & 0.920 & 0.610 & 0.331 & 0.371 
\\
TimeLLM & 0.847 & \textbf{0.565} & \underline{0.315} & 0.357 & 0.868 & 0.595 & 0.322 & 0.363 \\
CALF & 0.819 & 0.586 & 0.317 & 0.357 & 0.833 & \textbf{0.582} & \textbf{0.318} & \underline{0.359} \\
T-LLM & \textbf{0.738} & \underline{0.568} & \underline{0.315} & \underline{0.356} & 0.814 & \textbf{0.582} & \textbf{0.318} & \textbf{0.358} \\ \bottomrule
\end{tabular}}
\end{table}

\subsubsection{Zero-shot Learning}
Table~\ref{tab:zero-shot} reports zero-shot forecasting performance on the ETT benchmarks, where models trained on one subset are evaluated on another without further adaptation. Results are averaged across four prediction horizons $\{96, 192, 336, 720\}$. Compared to CALF, T-LLM achieves a consistent reduction in forecasting error across the evaluated transfer settings, with an overall error decrease of 2.2\%, reflecting improvements in both MSE and MAE. These results indicate that temporal supervision enables the LLM to internalize transferable forecasting behavior, leading to more robust generalization across related but distinct datasets.

% tab:complexity
\begin{table}[t]
\centering
\caption{Computational cost of LLM-based forecasters.}
\label{tab:complexity}
\resizebox{0.4\textwidth}{!}{\begin{tabular}{@{}cccccc@{}}
\toprule
 & T-LLM & CALF & Time-LLM & GPT4TS & UniTime \\ \midrule
FLOPs & \textbf{0.942G} & \textbf{0.942G} & 76.117G & 3.588G & 4.52G \\
Params & 0.101G & 0.18G & 0.135G & \textbf{0.083G} & 0.109G \\ \bottomrule
\end{tabular}}
\end{table}

\subsection{Efficiency Analysis}
As shown in Table~\ref{tab:complexity}, we report the parameter counts and inference FLOPs of LLM-based forecasting methods under the same experimental setting. T-LLM incurs the lowest computational cost among all compared methods, while maintaining competitive parameter efficiency. In contrast, LLM-based baselines such as Time-LLM involve additional prompt-generation procedures at inference time, resulting in substantially higher computational overhead. Overall, T-LLM achieves the best inference efficiency and the second-lowest parameter count, highlighting its favorable trade-off between model complexity and forecasting performance.

\subsection{Ablation Studies}

\begin{table}[t]
\centering
\caption{Ablation of loss components on ETTm1 and Weather.}
\label{tab:abl_loss}
\resizebox{0.44\textwidth}{!}{\begin{tabular}{cccccccc}
\toprule
\multirow{2}{*}{$\mathcal{L}_{\text{guide}}$} & \multirow{2}{*}{$\mathcal{L}_{\text{imit}}$} & \multirow{2}{*}{$\mathcal{L}_{\text{stud}}$} & \multirow{2}{*}{$\mathcal{L}_{\text{teach}}$} & \multicolumn{2}{c}{ETTm1} & \multicolumn{2}{c}{weather} \\ \cmidrule(lr){5-6} \cmidrule(lr){7-8} 
 &  &  &  & MSE & MAE & MSE & MAE \\ \hline
$\checkmark$ & - & $\checkmark$ & $\checkmark$ & 0.406 & 0.395 & 0.249 & 0.275 \\
- & $\checkmark$ & $\checkmark$ & $\checkmark$ & 0.397 & 0.391 & 0.250 & 0.273 \\
- & - & $\checkmark$ & $\checkmark$ & 0.402 & 0.395 & 0.253 & 0.279 \\
$\checkmark$ & $\checkmark$ & - & $\checkmark$ & 0.397 & 0.391 & 0.252 & 0.275 \\
$\checkmark$ & $\checkmark$ & $\checkmark$ & $\checkmark$ & \textbf{0.393} & \textbf{0.389} & \textbf{0.247} & \textbf{0.271} \\
\bottomrule
\end{tabular}}
\end{table}

\subsubsection{Loss Function Components}
Table~\ref{tab:abl_loss} examines the contribution of individual loss components in the reverse distillation framework on ETTm1 and Weather. Removing the student supervision loss $\mathcal{L}_{\text{stud}}$ leads to a consistent degradation in forecasting performance, resulting in an average error increase of of 1.40\% (MSE) and 0.95\% (MAE) across datasets. This indicates that direct supervision on the student prediction is necessary to regularize training and promote the student to catch up with teacher's performance. In addition, disabling the imitation loss $\mathcal{L}_{\text{imit}}$ or the guidance loss $\mathcal{L}_{\text{guide}}$ causes clear performance drops of 0.86\% and 1.75\%, respectively, despite retaining supervision losses. Removing both losses simultaneously further increases the error by 2.29\%. These results confirm that reverse distillation provides essential teacher-guided temporal information that cannot be replaced by prediction-level supervision alone.

\begin{figure}[t]
     \begin{subfigure}[t]{0.23\textwidth}
         \centering
         \includegraphics[width=\textwidth]{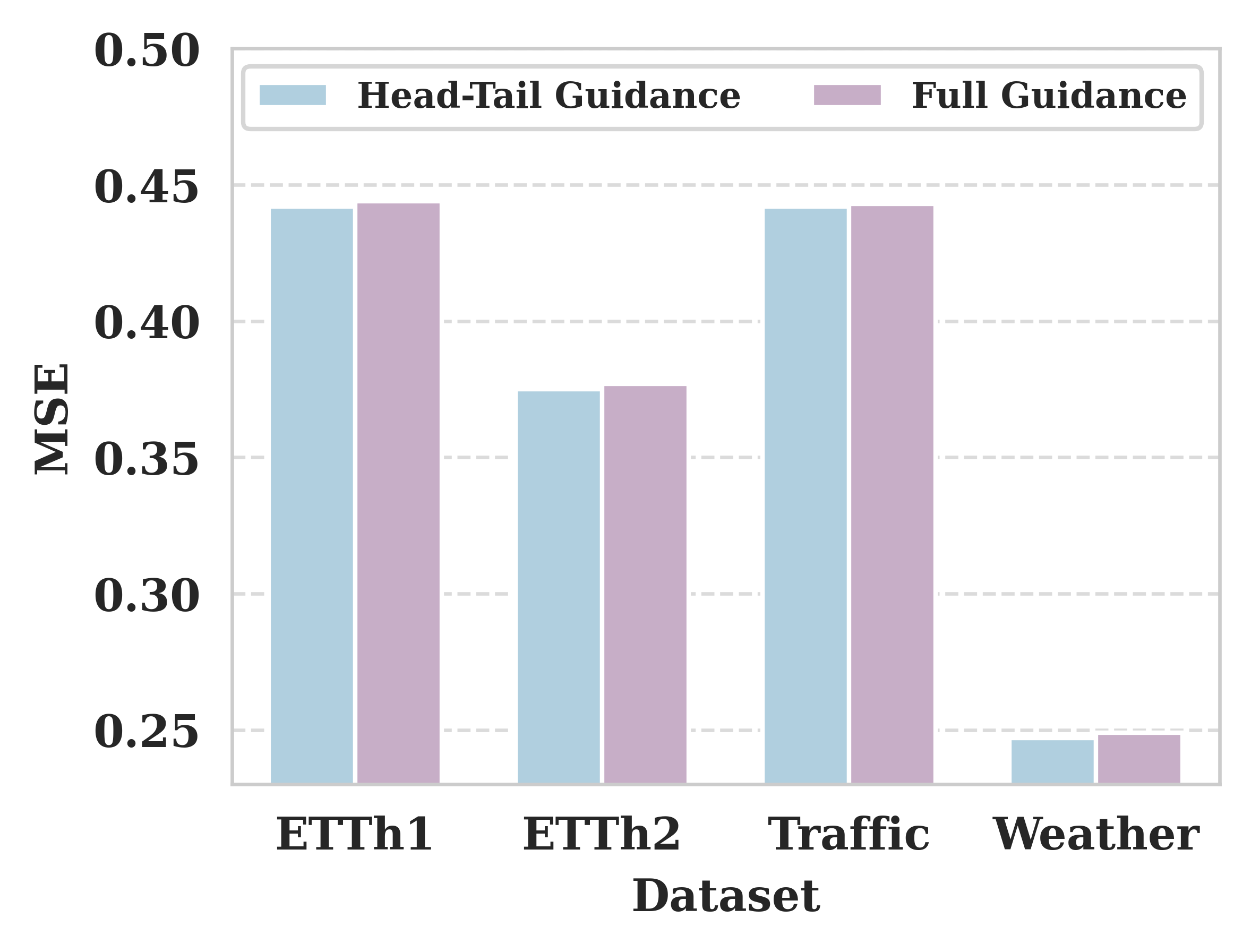}
         \caption{MSE}
         \label{fig:abl_headtail_MSE}
     \end{subfigure}
     % \hspace{16mm}
     \begin{subfigure}[t]{0.23\textwidth}
         \centering
         \includegraphics[width=\textwidth]{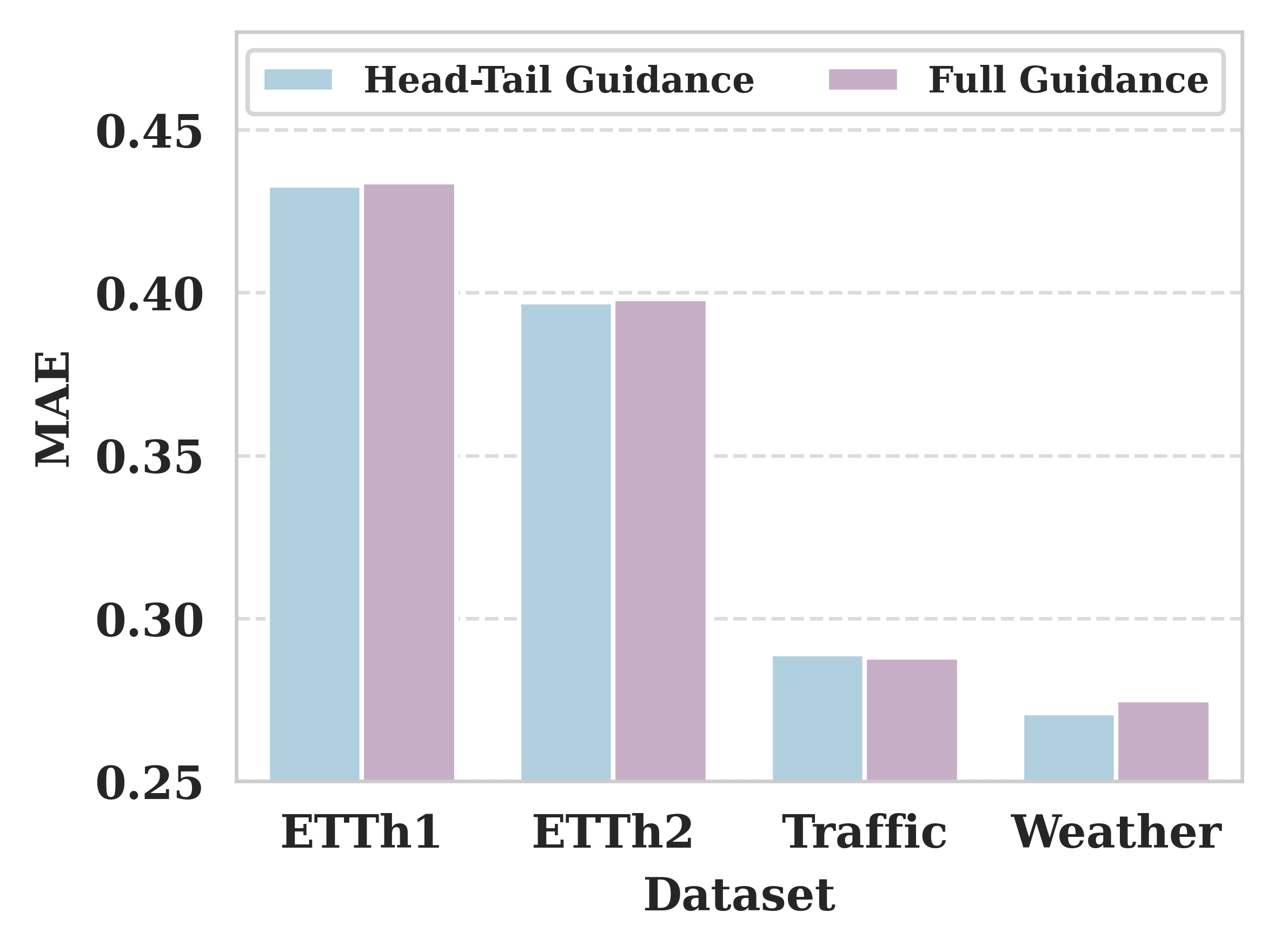}
         \caption{MAE}
         \label{fig:abl_headtail_MAE}
     \end{subfigure}
\caption{Ablation of head–tail guidance compared to full layer-wise guidance.}
\label{fig:abl_headtail}
\end{figure}

\subsubsection{Head–Tail Guidance Strategy}

We evaluate the effect of guidance granularity by comparing the proposed head–tail guidance strategy with full layer-wise guidance. As shown in Figure~\ref{fig:abl_headtail}, replacing head–tail guidance with full guidance results in only marginal changes in forecasting error, with differences generally within 1\% across datasets, indicating that enforcing guidance at every layer offers no meaningful accuracy advantage over selectively aligning early and late layers.

In contrast, full layer-wise guidance incurs substantially higher computational and parameter costs, requiring approximately $N/2$ times more guidance modules than head–tail guidance for an $N$-layer LLM. Given the high hidden dimensionality ($768$ in our setting), the additional projection and $\text{Temp-Spec}(\cdot)$ operations introduce non-trivial parameter and FLOPs overhead. Head–tail guidance therefore provides a more cost-effective design, achieving slightly better performance while significantly reducing guidance-related computation.

% Version 1
% We study the effect of alignment granularity by comparing the proposed head–tail alignment strategy against aligning all intermediate transformer layers between the LLM and the temporal branch. Table~X reports the relative error increase when replacing head–tail alignment with full layer-wise alignment. Across datasets, aligning all layers leads to an average error increase of 1.65\% (MSE) and 1.03\% (MAE), with a more pronounced degradation on the ETT datasets, where the average increase reaches 2.40\% (MSE) and 1.10\% (MAE).

% An exception is observed on the ECL dataset, where full alignment yields a slight improvement. Compared to ECL, the ETT datasets share similar characteristics, including a smaller number of channels and simpler inter-variable dependencies. In these settings, aligning all layers appears to over-regularize the LLM, constraining feature learning and leading to performance degradation. In contrast, restricting alignment to the input and output layers provides sufficient temporal grounding while preventing overfitting, resulting in more robust performance across datasets with limited channel complexity.

\begin{figure}[t]
     \begin{subfigure}[t]{0.23\textwidth}
         \centering
         \includegraphics[width=\textwidth]{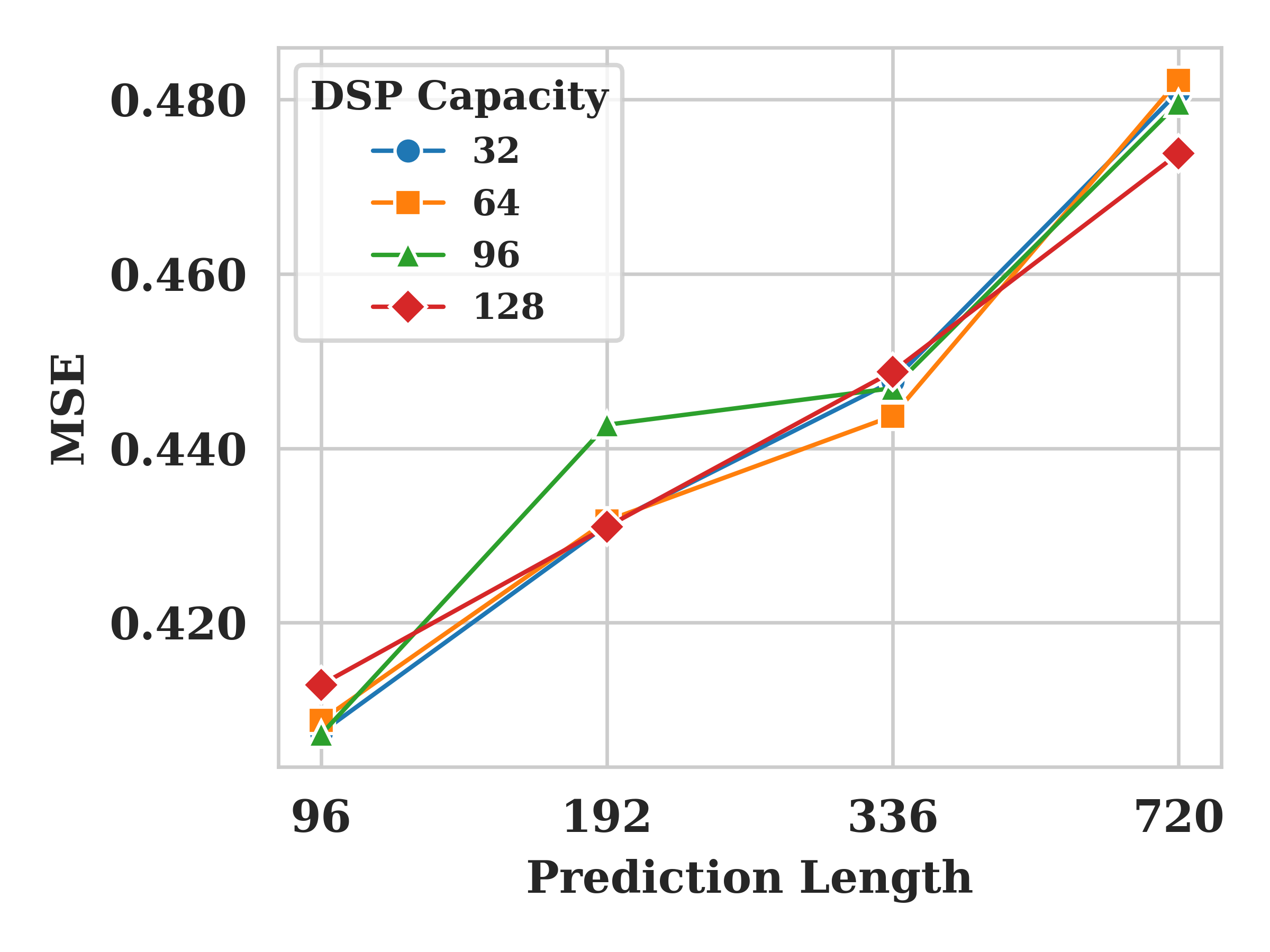}
         \caption{MSE}
         \label{fig:abl_DSP_MSE}
     \end{subfigure}
     % \hspace{16mm}
     \begin{subfigure}[t]{0.23\textwidth}
         \centering
         \includegraphics[width=\textwidth]{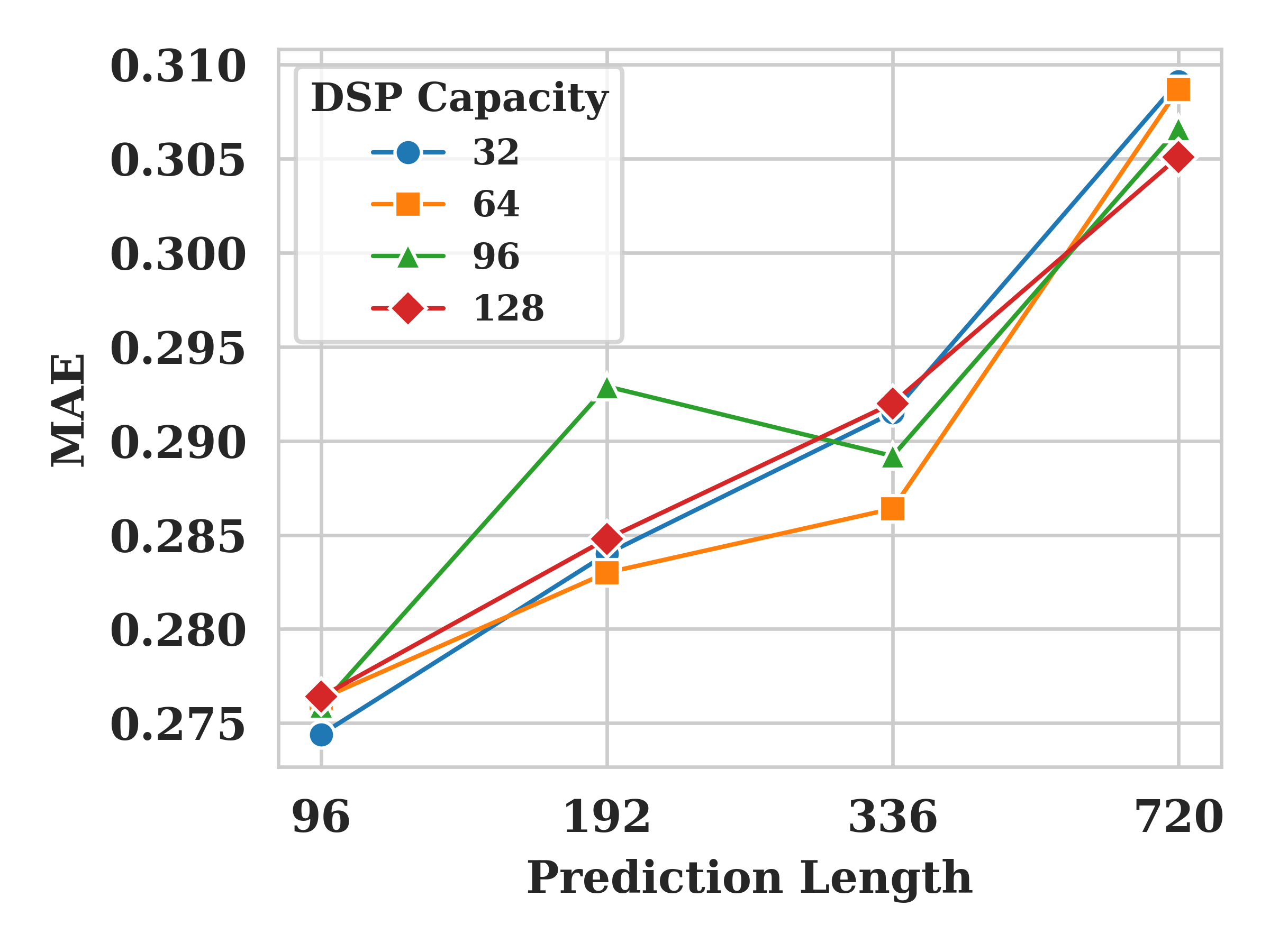}
         \caption{MAE}
         \label{fig:abl_DSP_MAE}
     \end{subfigure}
\caption{Ablation of DSP capacity choices ($d_{\text{red}}\in\{32,64,96,128\}$) across different prediction horizons.}
\label{fig:abl_DSP}
\end{figure}

\subsubsection{Horizon-Conditioned DSP Capacity}

Figure~\ref{fig:abl_DSP} evaluates the effect of the DSP hidden dimension under different prediction lengths on the Traffic dataset. As prediction length increases, forecasting error rises consistently for all configurations, reflecting the increasing difficulty of long-horizon forecasting. We compare four fixed capacity choices ($d_{\text{red}}\in\{32,64,96,128\}$) and observe that no single setting is uniformly optimal across horizons, which motivates our horizon-conditioned DSP design that selects an appropriate capacity based on the forecasting length. Importantly, the performance variation across capacity choices is small overall (within approximately 0.002 in MSE and 0.001 in MAE), indicating that DSP capacity selection is not overly sensitive to hyperparameters. This suggests that horizon-conditioned selection serves primarily as a robust design choice that recovers near-optimal performance across horizons without requiring careful manual tuning.

% tab: case study
\begin{table}[t]
\centering
\caption{Cross-epidemic zero-shot forecasting results on real-world influenza and COVID-19 datasets. }
\label{tab:casestudy}
\resizebox{0.5\textwidth}{!}{\begin{tabular}{@{}ccccccccc@{}}
\toprule
Model & \multicolumn{2}{c}{FLU → COVID Inc.} & \multicolumn{2}{c}{FLU → COVID Mot.} & \multicolumn{2}{c}{COVID Mot. → FLU} & \multicolumn{2}{c}{COVID Inc. → FLU} \\ \cmidrule(lr){2-3} \cmidrule(lr){4-5} \cmidrule(lr){6-7} \cmidrule(lr){8-9}
 & MSE & MAE & MSE & MAE & MSE & MAE & MSE & MAE \\ \midrule
GPT4TS & 1.199 & 0.711 & 0.803 & 0.472 & 2.657 & 0.653 & 3.602 & 0.742 \\
Time-LLM & 1.122 & 0.615 & 0.713 & 0.407 & \textbf{2.484} & \textbf{0.647} & 3.245 & 0.736 \\
CALF & \underline{1.004} & \underline{0.571} & \underline{0.662} & \underline{0.380} & \underline{2.655} & \underline{0.649} & \underline{2.927} & \underline{0.674} \\
T-LLM & \textbf{0.875} & \textbf{0.517} & \textbf{0.577} & \textbf{0.341} & 2.677 & 0.657 & \textbf{2.726} & \textbf{0.663} \\ \bottomrule
\end{tabular}}
\end{table}

\subsection{Case Study: Epidemiological Forecasting}

Infectious disease forecasting is a real-world setting characterized by limited and heterogeneous data, where models must often operate in a zero-shot or near–zero-shot regime. We use this case study to evaluate whether forecasting behavior learned through lightweight temporal supervision can generalize effectively under realistic data constraints. Table~\ref{tab:casestudy} reports cross-epidemic zero-shot forecasting results on real-world influenza and COVID-19 datasets. Models are trained on one epidemic (influenza or COVID-19) and evaluated on another without further adaptation, with results averaged over horizons $\{7,14,21,28\}$. T-LLM achieves the best performance in most transfer settings, particularly when forecasting COVID-19 incidence (Inc.) or mortality (Mot.) from influenza data, demonstrating improved cross-epidemic generalization under severe data scarcity.

\section{Conclusion}

This work addresses whether forecasting behavior can be taught to a general-purpose language model under the constraint that time series data cannot be scaled arbitrarily through pretraining. We propose T-LLM, a temporal distillation framework that transfers forecasting behavior from a lightweight temporal teacher to an LLM during training, while retaining the LLM as the sole inference-time model. Extensive experiments across benchmark datasets and real-world infectious disease forecasting tasks demonstrate that lightweight temporal supervision enables effective full-shot, few-shot, and zero-shot forecasting without reliance on large-scale time-series pretraining or inference-time temporal modules. Our results suggest that teaching forecasting behavior through structured temporal guidance provides a complementary and scalable alternative to purely scale-driven approaches for adapting LLMs to time-dependent domains.

\begin{acks}
Omitted for anonymity.
\end{acks}

\bibliographystyle{ACM-Reference-Format}
\bibliography{T-LLM}

\appendix

\section{Model Structure Details}
\subsection{Adaptive Spectral Block} \label{app-sec:asb_block}

Follow TSLAnet~\cite{eldeleTSLANetRethinkingTransformers2024}, We first transform temporal representation $\mathbf{E}_1$ into the frequency domain using a one-dimensional Fast Fourier Transform (FFT) applied along the temporal dimension:
\begin{equation}
    \mathbf{F} = \mathcal{F}(\mathbf{E}_1) \in \mathbb{C}^{C \times d_{\text{FFT}}},
    \quad
    d_{\text{FFT}} = \frac{d_{\text{model}}}{2} + 1.
\end{equation}
where $\mathcal{F}(\cdot)$ denotes the FFT operator, and $\mathbf{F}$ encodes spectral characteristics across channels. The transformed frequency dimension $d_{\text{FFT}}$ corresponds to the retained Fourier coefficients for real-valued inputs. Following standard spectral analysis, we compute the power spectrum to quantify the strength of each frequency component and use it to construct an adaptive frequency mask:
\begin{equation}
    \mathbf{P} = |\mathbf{F}|^2, \quad \mathbf{M} = \mathbb{I}(\mathbf{P} > \theta),
\end{equation}
where $\theta$ is a learnable threshold and $\mathbb{I}(\cdot)$ denotes an element-wise indicator function. This masking operation suppresses weak spectral components while preserving dominant frequencies. With the filtered representation, we further capture both global and locally filtered spectral information using two sets of learnable frequency-domain modulation coefficients:
\begin{equation}
    \mathbf{F}_G = \boldsymbol{\Gamma}_G \odot \mathbf{F}, \quad
    \mathbf{F}_L = \boldsymbol{\Gamma}_L \odot (\mathbf{F} \odot \mathbf{M}),
\end{equation}
where $\boldsymbol{\Gamma}_G$ and $\boldsymbol{\Gamma}_L$ denote global and local spectral modulation coefficients, and $\odot$ denotes element-wise multiplication. The resulting representations are integrated as:
\begin{equation}
    \mathbf{F}_{\text{spec}} = \mathbf{F}_G + \mathbf{F}_L.
\end{equation}

\subsection{Pooling} \label{app-sec:pooling}

According to T3Time~\cite{chowdhuryT3TimeTriModalTime2025}, we aggregate frequency-domain features via pooling. After dominant spectral projection, the reduced spectral representation is
$\mathbf{F}_{\text{spec}} \in \mathbb{R}^{C \times d_{\text{red}}}$. To enable attention-based aggregation across frequency bins, we first lift each frequency token to the model dimension:
\begin{equation}
    \mathbf{Z}_{\text{spec}}
    =
    \mathbf{F}_{\text{spec}} \mathbf{W}_f
    \in
    \mathbb{R}^{C \times d_{\text{red}} \times d_{\text{model}}},
\end{equation}
where $\mathbf{W}_f \in \mathbb{R}^{1 \times d_{\text{model}}}$ denotes a learnable projection shared across frequency responses. This produces $d_{\text{red}}$ frequency tokens per channel in the shared latent space. We then compute attention weights $\boldsymbol{\alpha} \in \mathbb{R}^{C \times d_{\text{red}} \times 1}$ over the frequency tokens using a two-layer perceptron $\text{MLP}(\cdot)$:
\begin{equation}
    \boldsymbol{\alpha}
    =
    \text{Softmax}\!\left(
    \text{MLP}(\mathbf{Z}_{\text{spec}})
    \right),
\end{equation}
and aggregate them via a weighted sum:
\begin{equation}
    \tilde{\mathbf{F}}
    =
    \sum_{l=1}^{d_{\text{red}}}
    \alpha_l \cdot \mathbf{Z}_{\text{spec},:,l}
    \in
    \mathbb{R}^{C \times d_{\text{model}}}.
\end{equation}

\subsection{Trend–Periodic Aggregation} \label{app-sec:gating}

To combine complementary temporal signals captured by the trend modeling and frequency modeling components, we introduce a Trend–Periodic Fusion Gate that adaptively balances their contributions using a continuous conditioning signal derived from the forecasting setup~\cite{chowdhuryT3TimeTriModalTime2025}. This module enables smooth interpolation between coarse trend dynamics and periodic temporal patterns within the temporal teacher branch.

Let $\tilde{\mathbf{H}}$ denote the trend-based temporal representation produced by the linear decomposition module, and let $\tilde{\mathbf{F}}$ denote the periodic temporal representation obtained from the frequency modeling pathway. We first pool $\tilde{\mathbf{H}}$ over the feature dimension and concatenate the result with a normalized forecast length to form the conditioning vector $\mathbf{g}_{\text{in}}$. The fusion weights are obtained as
\begin{equation}
    \mathbf{g} = \text{MLP}(\mathbf{g}_{\text{in}}),
    \quad \mathbf{g} \in \mathbb{R}^{C}.
\end{equation}

The fused temporal representation is obtained as a convex combination of the trend-based and periodic representations:
\begin{equation}
    \mathbf{E}_2
    =
    \mathbf{g} \odot \tilde{\mathbf{F}}
    +
    (1 - \mathbf{g}) \odot \tilde{\mathbf{H}},
\end{equation}
This fusion mechanism allows the temporal teacher to flexibly adjust the relative contribution of trend-based and periodic temporal information based on the forecasting configuration.

% Dataset Statistics Table
\begin{table*}[t!]
    \centering
    \caption{Dataset statistics are from~\cite{wuTimesNetTemporal2DVariation2023}. The dimension indicates the number of time series (i.e., channels), and the dataset size is organized in (training, validation, testing).}
    \label{tab:data_statistics}
    \resizebox{0.75\textwidth}{!}{\begin{tabular}{l|c|c|c|c|c}
    \toprule
    Dataset & Output Length & Dataset Size & Granularity & Dim. & Domain \\
    \midrule
    Weather & \{96, 192, 336, 720\} & (36456, 5175, 10444) & 10 min & 21 & Weather \\
    Traffic & \{96, 192, 336, 720\} & (11849, 1661, 3413) & 1 hour & 862 & Transportation \\
    Electricity & \{96, 192, 336, 720\} & (17981, 2537, 5165) & 1 hour & 321 & Electricity \\
    ILI & \{24, 36, 48, 60\} & (549, 74, 170) & 1 week & 7 & Illness \\
    ETTm1, ETTm2 &  \{96, 192, 336, 720\} & (34129, 11425, 11425) & 15 min & 7 & Temperature \\
    ETTh1, ETTh2 & \{96, 192, 336, 720\} & (8209, 2785, 2785) & 1 hour & 7 & Temperature \\
    \bottomrule
    \end{tabular}}
\end{table*}

% TODO: ADD dataset details about case study datasets
\section{Dataset Details}\label{app-sec:dset}

The detailed information for the dataset is presented below:
\begin{enumerate}
    % Weather
    \item The Weather dataset~\cite{wuTimesNetTemporal2DVariation2023} includes one-year records from 21 meteorological stations located in Germany, with a sampling rate of 10 minutes.
    % Traffic
    \item The traffic dataset~\cite{wuTimesNetTemporal2DVariation2023} describes the road occupancy rates. It contains the hourly data recorded by the 762 sensors of San Francisco freeways from 2015 to 2016.
    % Electricity
    \item The Electricity dataset~\cite{wuTimesNetTemporal2DVariation2023} comprises two-year records of electricity consumption from 321 customers, measured at a 1-hour sampling rate.
    % ILI
    \item The influenza-like illness (ILI) dataset~\cite{wuTimesNetTemporal2DVariation2023} contains records of patients experiencing severe influenza with complications.
    % ETT
    \item The Electricity Transformer Temperature (ETT; An indicator reflective of long-term electric power deployment)~\cite{zhouInformerEfficientTransformer2021} benchmark is comprised of two years of data, sourced from two counties in China. It comprises two hourly-level datasets (ETTh) and two 15-minute-level datasets (ETTm). Each entry within the ETT datasets includes six power load features and a target variable, termed “oil temperature”.
\end{enumerate}
Furthermore, the data statistics can be found in Table~\ref{tab:data_statistics}.

\section{Evaluation Metrics}

We evaluate most forecasting performance using two commonly used metrics: Mean Squared Error (MSE) and Mean Absolute Error (MAE). MSE and MAE measure the average prediction error across all forecast horizons, channels, and evaluation samples. MSE emphasizes larger deviations through squared errors, while MAE provides a linear and more robust error measure. Both metrics are computed over the full multivariate forecast window and averaged across samples.

Let $\hat{\mathbf{Y}}^{(i)} \in \mathbb{R}^{T \times C}$ denote the predicted future sequence and $\mathbf{Y}^{(i)} \in \mathbb{R}^{T \times C}$ the corresponding ground truth for sample i, where T is the prediction horizon and C is the number of channels. Given N evaluation samples, the Mean Squared Error (MSE) and Mean Absolute Error (MAE) are defined as:
\begin{align}
    \mathrm{MSE}
    &=
    \frac{1}{N T C}
    \sum_{i=1}^{N}
    \left\|
    \hat{\mathbf{Y}}^{(i)} - \mathbf{Y}^{(i)}
    \right\|_2^2, \\
    \mathrm{MAE}
    &=
    \frac{1}{N T C}
    \sum_{i=1}^{N}
    \left\|
    \hat{\mathbf{Y}}^{(i)} - \mathbf{Y}^{(i)}
    \right\|_1.
\end{align}

In addition to MSE and MAE, we report Symmetric Mean Absolute Percentage Error (SMAPE), Mean Absolute Scaled Error (MASE), and the Overall Weighted Average (OWA) for short-term forecasting on the M4 benchmark.

% Symmetric Mean Absolute Percentage Error (SMAPE)
SMAPE measures relative forecasting error while symmetrically normalizing by both prediction and ground truth magnitudes. SMAPE is defined as:
\begin{equation}
    \mathrm{SMAPE} = 
    \frac{200}{N T C}
    \sum_{i=1}^{N}
    \sum_{t=1}^{T}
    \sum_{c=1}^{C}
    \frac{
    \left|
    \hat{\mathbf{Y}}^{(i)}_{t,c} - \mathbf{Y}^{(i)}_{t,c}
    \right|
    }{
    \left|
    \hat{\mathbf{Y}}^{(i)}_{t,c}
    \right|
    +
    \left|
    \mathbf{Y}^{(i)}_{t,c}
    \right|
    },
\end{equation}
where zero denominators are handled by standard numerical safeguards. SMAPE is scale-independent and widely used for heterogeneous time series.

%Mean Absolute Scaled Error (MASE)
MASE evaluates forecast accuracy relative to a naive seasonal baseline, enabling fair comparison across series with different scales. For each sample $i$, MASE is computed as:
\begin{equation}
    \mathrm{MASE}=
    \frac{
    \frac{1}{T C}
    \left\|
    \hat{\mathbf{Y}}^{(i)} - \mathbf{Y}^{(i)}
    \right\|_1
    }{
    \frac{1}{(L-m) C}
    \sum_{t=m+1}^{L}
    \left\|
    \mathbf{X}^{(i)}_t - \mathbf{X}^{(i)}_{t-m}
    \right\|_1
    },
\end{equation}
where $m$ denotes the seasonal period and the denominator corresponds to the in-sample error of a seasonal naive predictor. Final MASE scores are averaged across samples.

% Overall Weighted Average (OWA)
The Overall Weighted Average (OWA) combines SMAPE and MASE by comparing a model’s performance against the Naive2 baseline. OWA is defined as:
\begin{equation}
\mathrm{OWA}
=
\frac{1}{2}
\left(
\frac{\mathrm{SMAPE}}{\mathrm{SMAPE}_{\text{Naive2}}}
+
\frac{\mathrm{MASE}}{\mathrm{MASE}_{\text{Naive2}}}
\right),
\end{equation}
where $\mathrm{SMAPE}_{\text{Naive2}}$ and $\mathrm{MASE}_{\text{Naive2}}$ denote the corresponding error metrics achieved by the Naive2 forecasting method. The Naive2 method is the standard benchmark used in the M4 competition. It first removes seasonality from the time series using a multiplicative seasonal component estimated from the training data, applies a naive forecast by repeating the last observed value on the deseasonalized series, and then reintroduces the seasonal component to generate the final prediction. By normalizing errors against Naive2, OWA enables fair comparison across datasets with different scales and seasonal characteristics.

\begin{figure*}[t]
    \centering
    \includegraphics[width=0.9\linewidth]{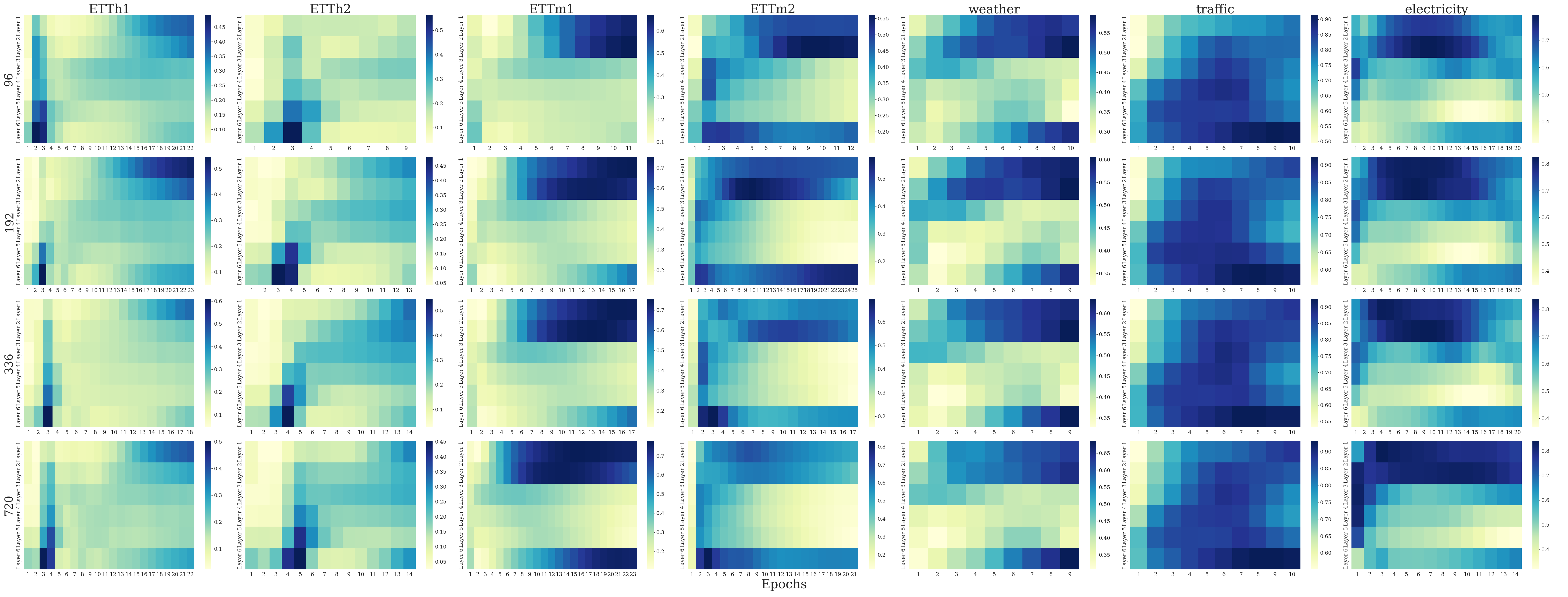}
    \caption{Visualization of attention centered kernel alignment (CKA) similarity across datasets. The y-axis denotes transformer layers, and the x-axis denotes training epochs.}
    \label{fig:attention_similarity}
\end{figure*}

\begin{figure*}[t]
     \begin{subfigure}[t]{0.9\textwidth}
         \centering
         \includegraphics[width=\textwidth]{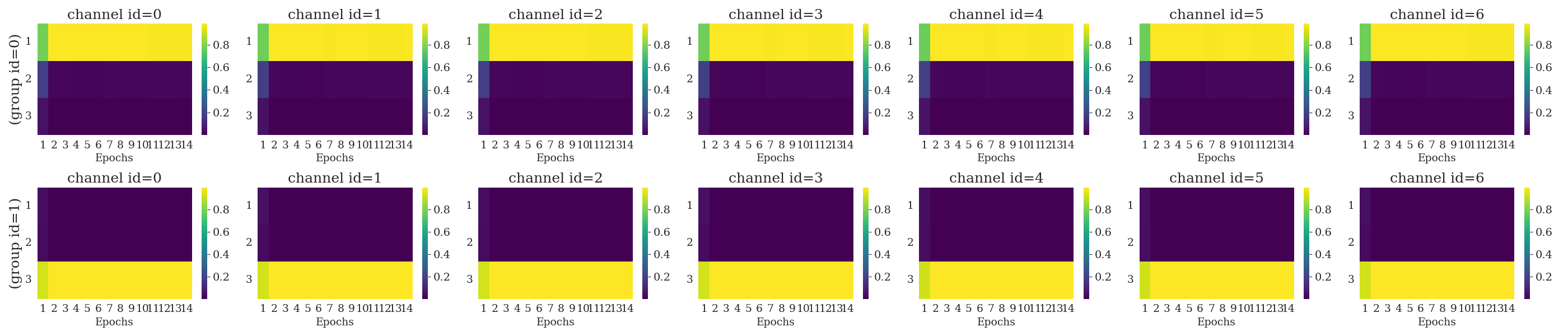}
         \caption{ETTh1, Forecasting Horizon: 96}
         \label{fig:att_weights_96}
     \end{subfigure}
     \hspace{16mm}
     \begin{subfigure}[t]{0.9\textwidth}
         \centering
         \includegraphics[width=\textwidth]{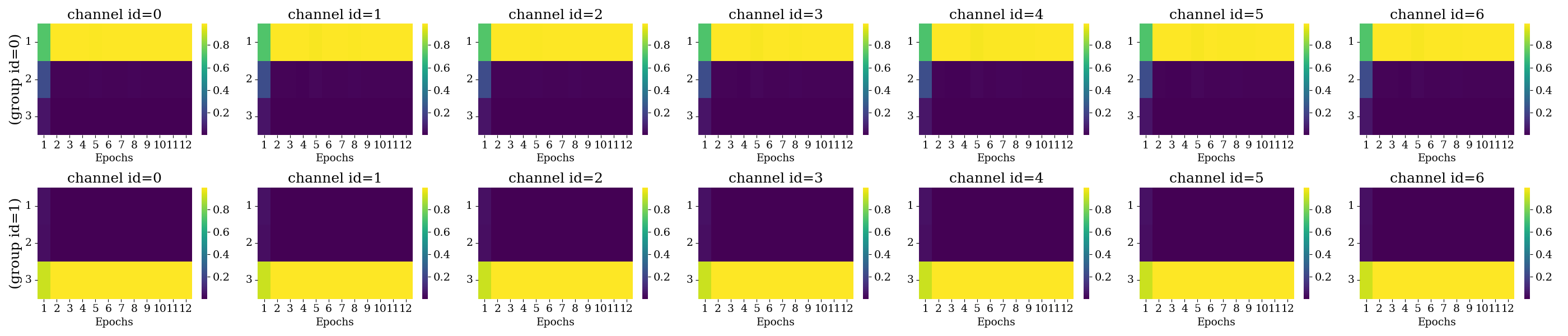}
         \caption{ETTh1, Forecasting Horizon: 720}
         \label{fig:att_weights_720}
     \end{subfigure}
\caption{Attention weights aggregated by layer groups, where each group consists of three attention layers. The y-axis denotes transformer layers, and the x-axis denotes training epochs.}
\label{fig:att_weights_ETTh1}
\end{figure*}

\section{Attention Analysis}\label{app-sec:att_decay}

% Prior work refers to TimeMixer++
We conduct a preliminary analysis of the CALF framework to examine the contribution of different transformer layers, motivating our use of head–tail guidance. Prior work suggests that higher centered kernel alignment (CKA) similarity indicates more consistent representations across layers, which is beneficial for forecasting tasks as it better captures trends and periodic patterns.

As shown in Figure~\ref{fig:attention_similarity}, we measure the similarity between the textual and temporal branches by computing the CKA similarity of attention matrices at corresponding layers. Across most datasets, middle layers (e.g., layers 2–5) exhibit consistently low CKA similarity and show little increase over training epochs, suggesting limited shared temporal representations. While datasets such as Traffic display relatively higher similarity in middle layers during early training, this similarity diminishes as training progresses, indicating that such guidance is not stable.

Figure~\ref{fig:att_weights_ETTh1} further reports aggregated attention weights grouped by layers, where each group consists of three consecutive attention layers. Under the head–tail guidance setting, the head groups consistently emphasize input-related attention, while the tail groups focus on output-related attention, with this pattern remaining stable across training epochs and forecasting horizons. Intermediate groups, in contrast, show less structured attention behavior.

Taken together, these observations suggest that strong and stable representation guidance is primarily concentrated in the first and last layers, which are most relevant for capturing input structure and producing forecasts. This empirical evidence supports our decision to adopt head–tail guidance instead of full layer-wise guidance, reducing unnecessary regularization and computational overhead without sacrificing forecasting performance.

% \section{Long Term Forecasting Full Results}

% tab:fewshot_full
\begin{table*}[]
\centering
\caption{Full Results of few-shot forecasting results on 10\% training data. The input sequence is set to 96 for all benchmark datasets.}
\label{tab:fewshot_full}
\resizebox{\textwidth}{!}{\begin{tabular}{@{}c|c|cc|cc|cc|cc|cc|cc|cc|cc|cc|cc|cc|cc@{}}
\toprule
\multirow{2}{*}{Dataset} & Model & \multicolumn{2}{c|}{T-LLM} & \multicolumn{2}{c|}{CALF} & \multicolumn{2}{c|}{Time-LLM} & \multicolumn{2}{c|}{GPT4TS} & \multicolumn{2}{c|}{iTransformer} & \multicolumn{2}{c|}{PatchTST} & \multicolumn{2}{c|}{Crossformer} & \multicolumn{2}{c|}{FEDFormer} & \multicolumn{2}{c|}{TimesNet} & \multicolumn{2}{c|}{MICN} & \multicolumn{2}{c|}{Dlinear} & \multicolumn{2}{c}{TiDE} \\ \cmidrule(l){2-26} 
 & Pred Len & MSE & MAE & MSE & MAE & MSE & MAE & MSE & MAE & MSE & MAE & MSE & MAE & MSE & MAE & MSE & MAE & MSE & MAE & MSE & MAE & MSE & MAE & MSE & MAE \\ \midrule
\multirow{5}{*}{ETTh1} & 96 & \textbf{0.396} & \underline{0.412} & 0.406 & 0.420 & 0.446 & 0.438 & 0.398 & \textbf{0.405} & 0.444 & 0.443 & 0.563 & 0.502 & 0.553 & 0.500 & 0.431 & 0.452 & 0.640 & 0.544 & \underline{0.398} & 0.428 & 0.540 & 0.491 & 0.802 & 0.602 \\
 & 192 & \underline{0.446} & \underline{0.439} & 0.453 & 0.444 & 0.497 & 0.465 & 0.448 & \textbf{0.433} & 0.499 & 0.474 & 0.611 & 0.525 & 0.585 & 0.543 & 0.482 & 0.477 & 0.633 & 0.545 & \textbf{0.435} & 0.452 & 0.571 & 0.511 & 0.877 & 0.635 \\
 & 336 & \textbf{0.486} & \underline{0.458} & \underline{0.496} & 0.467 & 0.539 & 0.493 & \underline{0.496} & \textbf{0.456} & 0.513 & 0.477 & 0.625 & 0.535 & 0.698 & 0.609 & 0.562 & 0.521 & 0.733 & 0.590 & 0.524 & 0.535 & 0.598 & 0.532 & 0.882 & 0.645 \\
 & 720 & \textbf{0.498} & \textbf{0.479} & 0.527 & 0.499 & 0.536 & 0.507 & \underline{0.504} & \underline{0.480} & 0.539 & 0.513 & 0.612 & 0.545 & 0.858 & 0.708 & 0.577 & 0.541 & 0.734 & 0.611 & 0.601 & 0.597 & 0.616 & 0.568 & 0.922 & 0.667 \\
 & Avg & \textbf{0.456} & \underline{0.447} & 0.470 & 0.457 & 0.504 & 0.476 & \underline{0.461} & \textbf{0.443} & 0.499 & 0.476 & 0.603 & 0.527 & 0.673 & 0.590 & 0.513 & 0.498 & 0.685 & 0.572 & 0.489 & 0.503 & 0.581 & 0.526 & 0.871 & 0.637 \\ \midrule
\multirow{5}{*}{ETTh2} & 96 & \textbf{0.296} & \textbf{0.343} & 0.305 & 0.354 & 0.307 & 0.353 & \underline{0.303} & \underline{0.353} & 0.339 & 0.379 & 0.348 & 0.385 & 0.532 & 0.515 & 0.339 & 0.383 & 0.376 & 0.406 & 0.355 & 0.402 & 0.393 & 0.438 & 0.536 & 0.475 \\
 & 192 & \textbf{0.377} & \textbf{0.392} & 0.383 & \underline{0.394} & 0.395 & 0.406 & \underline{0.381} & 0.401 & 0.432 & 0.433 & 0.432 & 0.432 & 0.915 & 0.696 & 0.417 & 0.430 & 0.458 & 0.453 & 0.505 & 0.489 & 0.473 & 0.481 & 0.575 & 0.499 \\
 & 336 & 0.429 & \textbf{0.430} & 0.427 & \textbf{0.430} & 0.432 & 0.436 & \textbf{0.420} & \underline{0.432} & 0.470 & 0.464 & \underline{0.423} & 0.439 & 1.018 & 0.762 & 0.459 & 0.469 & 0.492 & 0.482 & 0.665 & 0.588 & 0.531 & 0.509 & 0.600 & 0.523 \\
 & 720 & \textbf{0.418} & \textbf{0.438} & \underline{0.422} & \underline{0.440} & 0.436 & 0.450 & 0.435 & 0.451 & 0.472 & 0.473 & 0.444 & 0.455 & 1.332 & 0.889 & 0.479 & 0.487 & 0.514 & 0.494 & 0.894 & 0.696 & 0.516 & 0.520 & 0.584 & 0.520 \\
 & Avg & \textbf{0.380} & \textbf{0.401} & \underline{0.384} & \underline{0.405} & 0.392 & 0.411 & 0.385 & 0.409 & 0.428 & 0.437 & 0.411 & 0.428 & 0.949 & 0.716 & 0.423 & 0.442 & 0.460 & 0.458 & 0.605 & 0.544 & 0.478 & 0.487 & 0.574 & 0.504 \\ \midrule
\multirow{5}{*}{ETTm1} & 96 & \underline{0.336} & \textbf{0.359} & 0.353 & 0.376 & 0.428 & 0.415 & 0.349 & 0.372 & 0.375 & 0.392 & \textbf{0.325} & \underline{0.361} & 0.355 & 0.390 & 0.407 & 0.434 & 0.356 & 0.389 & 0.351 & 0.398 & 0.402 & 0.411 & 0.376 & 0.393 \\
 & 192 & 0.378 & \textbf{0.382} & 0.396 & 0.398 & 0.457 & 0.430 & 0.388 & 0.391 & 0.426 & 0.418 & \underline{0.371} & \underline{0.387} & 0.395 & 0.413 & 0.443 & 0.456 & 0.411 & 0.419 & \textbf{0.368} & 0.411 & 0.430 & 0.426 & 0.412 & 0.410 \\
 & 336 & 0.415 & \textbf{0.404} & 0.420 & 0.413 & 0.487 & 0.446 & 0.422 & \textbf{0.412} & 0.458 & 0.438 & \textbf{0.395} & \textbf{0.404} & 0.701 & 0.568 & 0.468 & 0.473 & 0.479 & 0.456 & \underline{0.396} & 0.429 & 0.455 & 0.442 & 0.440 & 0.427 \\
 & 720 & 0.491 & 0.451 & 0.498 & 0.455 & 0.531 & 0.473 & 0.478 & \underline{0.442} & 0.530 & 0.478 & \underline{0.461} & \textbf{0.441} & 0.768 & 0.610 & 0.557 & 0.509 & 0.519 & 0.476 & \textbf{0.438} & 0.458 & 0.508 & 0.475 & 0.502 & 0.462 \\
 & Avg & \underline{0.405} & \underline{0.399} & 0.417 & 0.410 & 0.476 & 0.441 & 0.409 & 0.404 & 0.447 & 0.432 & \textbf{0.388} & \textbf{0.398} & 0.555 & 0.495 & 0.469 & 0.468 & 0.441 & 0.435 & \textbf{0.388} & 0.424 & 0.449 & 0.439 & 0.432 & 0.423 \\ \midrule
\multirow{5}{*}{ETTm2} & 96 & 0.180 & \underline{0.260} & \underline{0.179} & \textbf{0.259} & 0.193 & 0.277 & 0.180 & 0.263 & 0.192 & 0.277 & \textbf{0.178} & \underline{0.260} & 0.286 & 0.360 & 0.206 & 0.289 & 0.188 & 0.270 & 0.221 & 0.313 & 0.228 & 0.331 & 0.191 & 0.274 \\
 & 192 & 0.246 & \textbf{0.302} & 0.247 & 0.304 & 0.260 & 0.321 & \underline{0.245} & 0.305 & 0.256 & 0.316 & \textbf{0.244} & \underline{0.304} & 0.343 & 0.398 & 0.266 & 0.324 & 0.287 & 0.345 & 0.282 & 0.354 & 0.303 & 0.381 & 0.256 & 0.314 \\
 & 336 & 0.308 & \textbf{0.341} & 0.307 & \underline{0.342} & 0.319 & 0.355 & \underline{0.306} & 0.345 & 0.319 & 0.355 & \textbf{0.305} & 0.343 & 0.641 & 0.566 & 0.328 & 0.363 & 0.314 & 0.348 & 0.374 & 0.414 & 0.388 & 0.436 & 0.316 & 0.351 \\
 & 720 & \underline{0.404} & \textbf{0.397} & 0.406 & \underline{0.398} & 0.418 & 0.409 & 0.407 & 0.402 & 0.418 & 0.408 & \textbf{0.403} & 0.399 & 0.878 & 0.674 & 0.423 & 0.420 & 0.442 & 0.425 & 0.529 & 0.502 & 0.544 & 0.522 & 0.416 & 0.405 \\
 & Avg & 0.285 & \textbf{0.325} & 0.285 & \textbf{0.325} & 0.298 & 0.341 & \underline{0.284} & 0.329 & 0.297 & 0.339 & \textbf{0.283} & \underline{0.326} & 0.537 & 0.500 & 0.306 & 0.349 & 0.308 & 0.347 & 0.351 & 0.396 & 0.366 & 0.417 & 0.294 & 0.336 \\ 
 \midrule
 \multicolumn{2}{c|}{\textbf{1st Count}} &\textbf{8}	 &\textbf{13} &\textbf{0}	&\textbf{3}	&\textbf{0}	&\textbf{0}	&\textbf{1}	&\textbf{4}	&\textbf{0}	&\textbf{0}	&\textbf{8}	&\textbf{3}	&\textbf{0}	&\textbf{0}	&\textbf{0}	&\textbf{0}	&\textbf{0}	&\textbf{0}	&\textbf{3}	&\textbf{0}	&\textbf{0}	&\textbf{0}	&\textbf{0}	&\textbf{0} \\
 \bottomrule
\end{tabular}}
\end{table*}

\section{Few-shot Forecasting Full Results}
The complete results for the few-shot forecasting tasks of our model are presented in Tables ~\ref{tab:fewshot_full}. For all benchmark datasets, the input sequence length is fixed at $96$, and the prediction sequence lengths are set to $\{96, 192, 336, 720\}$. In the 10\% few-shot scenario, our model achieves state-of-the-art (SOTA) performance in 21 out of 40 cases across five diverse time series benchmarks. 

% tab:full_casestudy
\begin{table*}[t]
\centering
\caption{Full results of cross-epidemic zero-shot forecasting results on real-world influenza and COVID-19 datasets over different test ratios. The input length is $28$, and forecasting horizons are $\{7, 14, 21, 28\}$. Reported results are averaged over these horizons.}
\label{tab:casestudy_full}
\resizebox{0.75\textwidth}{!}{\begin{tabular}{@{}cccccccccc@{}}
\toprule
\multirow{2}{*}{Test Ratio} & \multirow{2}{*}{Model} & \multicolumn{2}{c}{FLU → COVID Inc.} & \multicolumn{2}{c}{FLU → COVID Mot.} & \multicolumn{2}{c}{COVID Mot. → FLU} & \multicolumn{2}{c}{COVID Inc. → FLU} \\ \cmidrule(l){3-10} 
 &  & MSE & MAE & MSE & MAE & MSE & MAE & MSE & MAE \\ \midrule
\multirow{4}{*}{0.2} & GPT4TS & 1.199 & 0.711 & 0.803 & 0.472 & 2.657 & 0.653 & 3.602 & 0.742 \\
 & Time-LLM & 1.122 & 0.615 & 0.713 & 0.407 & \textbf{2.484} & \textbf{0.647} & 3.245 & 0.736 \\
 & CALF & \underline{1.004} & \underline{0.571} & \underline{0.662} & \underline{0.380} & \underline{2.655} & \underline{0.649} & \underline{2.927} & \underline{0.674} \\
 & T-LLM & \textbf{0.875} & \textbf{0.517} & \textbf{0.577} & \textbf{0.341} & 2.677 & 0.657 & \textbf{2.726} & \textbf{0.663} \\ \midrule
\multirow{4}{*}{0.4} & GPT4TS & 20.496 & 2.236 & 3.273 & 0.933 & 1.545 & 0.499 & 2.056 & 0.568 \\
 & Time-LLM & 18.883 & 1.989 & 2.932 & 0.805 & \textbf{1.440} & \textbf{0.493} & 1.714 & 0.541 \\
 & CALF & \textbf{16.029} & \textbf{1.742} & \underline{2.705} & \underline{0.769} & \underline{1.541} & \underline{0.494} & \underline{1.703} & \underline{0.516} \\
 & T-LLM & \underline{16.162} & \underline{1.758} & \textbf{2.425} & \textbf{0.707} & 1.556 & 0.502 & \textbf{1.591} & \textbf{0.511} \\ \midrule
\multirow{4}{*}{0.6} & GPT4TS & 43.988 & 2.748 & 20.204 & 1.620 & 1.294 & 0.530 & 1.683 & 0.591 \\
 & Time-LLM & 40.385 & 2.467 & 18.138 & 1.413 & \textbf{1.208} & \underline{0.525} & 1.439 & 0.572 \\
 & CALF & \textbf{34.961} & \textbf{2.192} & \underline{16.810} & \underline{1.351} & \underline{1.287} & \textbf{0.523} & \underline{1.417} & \underline{0.542} \\
 & T-LLM & \underline{35.113} & \underline{2.217} & \textbf{15.175} & \textbf{1.245} & 1.303 & 0.532 & \textbf{1.331} & \textbf{0.539} \\ \midrule
\multirow{4}{*}{0.8} & GPT4TS & 1065.213 & 13.924 & 91.535 & 3.869 & 4.156 & 0.685 & \underline{4.305} & 0.771 \\
 & Time-LLM & 987.416 & 12.567 & 82.599 & 3.397 & \underline{4.059} & \textbf{0.680} & 4.639 & 0.742 \\
 & CALF & \textbf{856.843} & \textbf{11.254} & \underline{76.918} & \underline{3.267} & 4.135 & 0.675 & 4.368 & \underline{0.701} \\
 & T-LLM & \underline{873.546} & \underline{11.437} & \textbf{69.349} & \textbf{3.039} & \textbf{4.056} & \underline{0.683} & \textbf{4.248} & \textbf{0.698} \\ \bottomrule
\end{tabular}}
\end{table*}

\section{Case Study Full Results}

Table~\ref{tab:casestudy_full} extends the cross-epidemic zero-shot forecasting results reported in the main text by evaluating performance under different test ratios on real-world influenza and COVID-19 datasets. The test ratio controls the fraction of data reserved for evaluation, with larger ratios corresponding to more challenging forecasting settings, as a greater portion of the epidemic trajectory must be predicted. This evaluation is motivated by a practical consideration in epidemiological forecasting: while zero-shot forecasting assumes no access to target-domain training data, using a fixed test split, as is common in benchmark settings, does not fully capture the difficulty of early-stage outbreak forecasting, where limited observations are available and uncertainty is highest.

By varying the test ratio, we simulate increasingly difficult forecasting regimes that place greater emphasis on early epidemic dynamics. Across a wide range of test ratios, T-LLM consistently achieves competitive or superior performance compared to existing LLM-based baselines in most transfer directions. These results further support the robustness of temporal distillation for cross-epidemic zero-shot forecasting under realistic and challenging evaluation conditions.

\end{document}